\documentclass{article} % For LaTeX2e
\usepackage{colm2024_conference}

\usepackage{graphicx}
\usepackage{microtype}
\usepackage{hyperref}
\usepackage{url}
\usepackage{booktabs}

\usepackage{amsmath} 
\usepackage{wrapfig}
\usepackage{algorithm}
\usepackage{algpseudocode}
\usepackage{longtable}
\usepackage{pgfplots}
\usepgfplotslibrary{groupplots}
\pgfplotsset{compat=newest}
\usepackage{xcolor}
\definecolor{darkblue}{rgb}{0, 0, 0.5}
\definecolor{lightgreen}{HTML}{3CB371} % Define color using HTML notation
\definecolor{darkgreen}{HTML}{2F4F4F} % Define color using HTML notation

\definecolor{lightgreen}{HTML}{64D2CE} % Define color using HTML notation
\definecolor{darkgreen}{HTML}{9AADD2} % Define color using HTML notation

\definecolor{purple}{HTML}{F54A8A} % Define color using HTML notation
\definecolor{lightblue}{HTML}{87CEEB}
\definecolor{darkblue}{HTML}{24AADD}
\definecolor{lightred}{HTML}{FA8072}

\hypersetup{colorlinks=true, citecolor=darkblue, linkcolor=darkblue, urlcolor=darkblue}

\vspace{-2cm}

\title{

    Benchmarking Benchmark Leakage in Large Language \\ Models
}

\colmfinalcopy

\newcommand*\samethanks[1][\value{footnote}]{\footnotemark[#1]}

\author{
Ruijie Xu\textsuperscript{\textrm{1,3}}\thanks{~~Co-first authors.}\space\space\space\space\space\space
Zengzhi Wang\textsuperscript{\textrm{1,3}}\samethanks\space\space\space\space\space\space
Run-Ze Fan\textsuperscript{\textrm{1,3}}\samethanks\space\space\space\space\space\space
Pengfei Liu\textsuperscript{\textrm{1,2,3}}\thanks{~~Corresponding author} \\
\textsuperscript{1}Shanghai Jiao Tong University \
\textsuperscript{2}Shanghai Artificial Intelligence Laboratory \\
\textsuperscript{3}Generative AI Research Lab (GAIR) \\
\texttt{\{hsu.rayjay,zzwang.nlp\}@gmail.com} \quad runze.fan@icloud.com \quad pengfei@sjtu.edu.cn
}

\begin{document}

\maketitle

\begin{figure}[h]
\centering
\vspace{-1.18cm}
\begin{tikzpicture}
  \begin{groupplot}[
    group style={
      group size=2 by 1, % 2 plots in a row, 1 plot in a column
      ylabels at=edge left,
      horizontal sep=2.3cm % Separation between plots
    },
    xbar,
    width=.43\textwidth,
    height=15cm,
    ytick=data,
    ytick pos = left,
    nodes near coords,
    nodes near coords align={horizontal},
    % nodes near coords style={font=\tiny},
    nodes near coords style={
        /pgf/number format/.cd, 
        fixed,
        fixed zerofill, 
        precision=2, 
        /tikz/.cd, 
        font=\fontsize{4}{10}\selectfont
    },
    xmin=-17,xmax=170,
    xlabel={$\delta_\text{train-test}$ (\%)},
    ylabel = {Models},
    xlabel style = {font=\footnotesize},
    ylabel style = {font=\tiny},
    ylabel shift = -4 pt,
    tick label style = {font=\tiny},
    title style = {font = \small},
    title style={yshift=-4pt},
    enlarge y limits=0.05,
    y dir=reverse,
    legend style={at={(0.5,-0.20)},
      anchor=north,legend columns=-1},
    ]
  ]
  \nextgroupplot[title={GSM8K},  xmin=-15, xmax=90, symbolic y coords={Aquila2-7B, InternLM2-20B, InternLM2-7B, Aquila2-34B, Qwen-14B,    Qwen-7B,    ChatGLM2-6B,     Qwen-1.8B,  Baichuan2-13B-Base, InternLM-20B, Orca-2-7B,     Yi-34B,     Yi-6B, Aquila-7B,     Phi-2,  InternLM-7B,     ChatGLM3-6B,     LLaMA2-7B,    Phi-1.5,     Baichuan-7B,    Mistral-7B-v0.1,     LLaMA-7B, Grok-1,    Gemma-2B,     Gemma-7B,  Llama-3-8B,     Baichuan2-7B-Base,    Baichuan-13B-Base,    InternLM2-20B-Base,     DeepSeekMath-7B,     InternLM2-7B-Base, 
    }  
  ]
  %  ppl
  \addplot+[ darkblue, fill=lightblue, bar width=4pt, bar shift = 2pt] coordinates {
    (75.25,Aquila2-7B)
    (67.77,InternLM2-20B)
    (64.75,InternLM2-7B)
    (39.78,Aquila2-34B)
    (26.94,Qwen-14B)
    ( 25.81,Qwen-7B)
    (28.76,ChatGLM2-6B)
    ( 10.30,Qwen-1.8B)
    ( 11.82,Baichuan2-13B-Base)
    ( 16.49,InternLM-20B)
    (17.10,Orca-2-7B)
    (10.19,Yi-34B)
    (5.30,Yi-6B)
    (9.80,Aquila-7B)
    (-2.03,Phi-2)
    (4.13,InternLM-7B)
    (5.43,ChatGLM3-6B)
    (-0.29,LLaMA2-7B)
    ( -1.74,Phi-1.5)
    ( -1.08,Baichuan-7B)
    ( -0.63,Mistral-7B-v0.1)
    (-0.85,LLaMA-7B)
    (-0.38,Grok-1)
    ( -1.48,Gemma-2B)
    (-0.70,Gemma-7B)
    ( -0.87,Llama-3-8B)
    ( -1.23,Baichuan2-7B-Base)
    ( -0.91,Baichuan-13B-Base)
    ( -1.16,InternLM2-20B-Base)
    ( -0.97,DeepSeekMath-7B)
    ( -0.83,InternLM2-7B-Base)
  };
  % 5-gram   , bar shift=-0.3pt
  \addplot+[ purple, fill=purple!30!white, bar width=4pt, bar shift = -2.3pt] coordinates {
    ( 23.24,Aquila2-7B)
    ( 29.11,InternLM2-20B)
    (30.12,InternLM2-7B)
    ( 23.12,Aquila2-34B)
    ( 35.16,Qwen-14B)
    (35.54,Qwen-7B)
    (13.65,ChatGLM2-6B)
    (22.87,Qwen-1.8B)
    (17.85,Baichuan2-13B-Base)
    (10.68,InternLM-20B)
    (6.45,Orca-2-7B)
    (11.55,Yi-34B)
    ( 9.05,Yi-6B)
    ( 2.57,Aquila-7B)
    (10.60,Phi-2)
    ( 3.86,InternLM-7B)
    ( 1.57,ChatGLM3-6B)
    ( 3.82,LLaMA2-7B)
    ( 4.46,Phi-1.5)
    ( 1.66,Baichuan-7B)
    ( 0.96,Mistral-7B-v0.1)
    ( 1.07,LLaMA-7B)
    (-0.53,Grok-1)
    ( -0.57,Gemma-2B)
    (-1.91,Gemma-7B)
    ( -2.13,Llama-3-8B)
    ( -2.06,Baichuan2-7B-Base)
    ( -2.58,Baichuan-13B-Base)
    (-2.42,InternLM2-20B-Base)
    ( -2.64,DeepSeekMath-7B)
    ( -3.21,InternLM2-7B-Base)
  };

  \nextgroupplot[title={MATH}, legend to name=grouplegend, xmin=-15, xmax=170,
    symbolic y coords={Aquila2-7B, InternLM2-20B, Aquila2-34B, InternLM2-7B, Qwen-14B,    Qwen-7B,   Qwen-1.8B, ChatGLM3-6B, InternLM-7B, InternLM-20B, Gemma-2B,  Orca-2-7B,  Llama-3-8B,   Gemma-7B,    Yi-6B,  Yi-34B, Grok-1,   Phi-2,  LLaMA-7B,  DeepSeekMath-7B, ChatGLM2-6B, InternLM2-7B-Base, Phi-1.5, LLaMA2-7B,  Baichuan-7B, InternLM2-20B-Base,     Baichuan2-7B-Base,    Mistral-7B-v0.1,   Baichuan-13B-Base,   Aquila-7B, Baichuan2-13B-Base,  
    }  
  ]
   %  ppl
  \addplot+[ darkblue, fill=lightblue, bar width=4pt, bar shift = 2pt] coordinates {
    (158.76,Aquila2-7B)
    (72.44,InternLM2-20B)
    (70.23,Aquila2-34B)
    (45.22,InternLM2-7B)
    (20.34,Qwen-14B)
    (17.60,Qwen-7B)
    (9.79,Qwen-1.8B)
    (9.69,ChatGLM3-6B)
    (6.05,InternLM-7B)
    (4.85,InternLM-20B)
    (1.05,Gemma-2B)
    (1.31,Orca-2-7B)
    (-0.01,Llama-3-8B)
    (-0.63,Gemma-7B)
    (-0.55,Yi-6B)
    (-0.60,Yi-34B)
    (1.39,Grok-1)
    (-0.41,Phi-2)
    (-0.88,LLaMA-7B)
    (-0.64,DeepSeekMath-7B)
    (-0.92,ChatGLM2-6B)
    (-0.20,InternLM2-7B-Base)
    (-0.43,Phi-1.5)
    (-0.18,LLaMA2-7B)
    (-0.09,Baichuan-7B)
    (-0.61,InternLM2-20B-Base)
    (-0.13,Baichuan2-7B-Base)
    (-0.62,Mistral-7B-v0.1)
    (-0.53,Baichuan-13B-Base)
    (-0.48,Aquila-7B)
    (-0.56,Baichuan2-13B-Base)
  };
  %  5-gram
  \addplot+[ purple, fill=purple!30!white, bar width=4pt, bar shift = -2.3pt] coordinates {
    (15.24,Aquila2-7B)
    (20.48,InternLM2-20B)
    (18.50,Aquila2-34B)
    (17.67,InternLM2-7B)
    (3.15,Qwen-14B)
    (4.20,Qwen-7B)
    (9.94,Qwen-1.8B)
    (5.33,ChatGLM3-6B)
    (4.11,InternLM-7B)
    (0.71,InternLM-20B)
    (2.66,Gemma-2B)
    (2.24,Orca-2-7B)
    (3.06,Llama-3-8B)
    (3.54,Gemma-7B)
    (3.05,Yi-6B)
    (2.81,Yi-34B)
    (0.36,Grok-1)
    (2.07,Phi-2)
    (2.28,LLaMA-7B)
    (1.34,DeepSeekMath-7B)
    (1.57,ChatGLM2-6B)
    (0.58,InternLM2-7B-Base)
    (0.78,Phi-1.5)
    (0.09,LLaMA2-7B)
    (0.15,Baichuan-7B)
    (0.46,InternLM2-20B-Base)
    (-0.58,Baichuan2-7B-Base)
    (-0.49,Mistral-7B-v0.1)
    (-0.61,Baichuan-13B-Base)
    (-2.10,Aquila-7B)
    (-2.40,Baichuan2-13B-Base)
  };

  \end{groupplot}

  \begin{axis}[
        hide axis, 
        xmin=0, xmax=1, ymin=0, ymax=0.1, 
        legend columns=1, 
        legend style={
            at={(1.45, 0.21)}, 
            anchor=north, 
            font=\tiny, 
            /tikz/every even column/.append style={column sep=10pt}
        }
    ]
    
    \addlegendimage{xbar,xbar legend, fill=lightblue,draw=darkblue}
    \addlegendentry{PPL}
    \addlegendimage{xbar,xbar legend, fill=purple!30!white,draw=purple}
    \addlegendentry{N-gram Accuracy}
\end{axis}
\end{tikzpicture}
\vspace{-0.75cm}
\caption{The relative possibility that various models conduct verbatim training on the training set of a benchmark over test set to enhance capabilities (measured based on PPL and N-gram Accuracy). Models exhibiting near-zero possibilities suggest either the absence of training and test split or the use of both splits in the training process. \textbf{This metric does not imply cheating, but rather indicates the potential use of the benchmark data during the (pre-)training phase; while using benchmarks to enhance capabilities is acceptable, the lack of relevant documentation can reduce transparency, potentially resulting in unfair comparisons and hindering the field's healthy development.}}
\label{}
\end{figure}
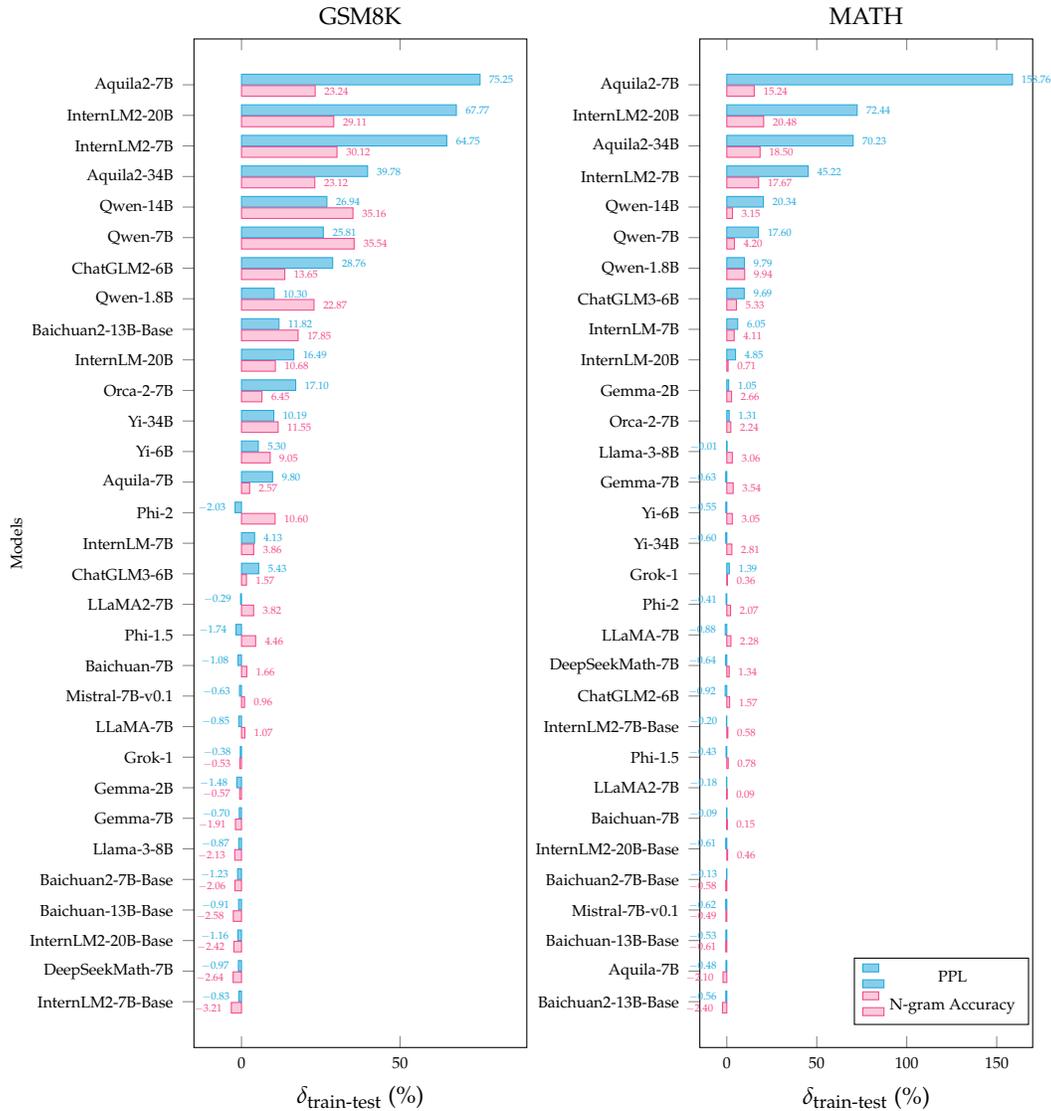

\begin{abstract}

Amid the expanding use of pre-training data, the phenomenon of benchmark dataset leakage has become increasingly prominent, exacerbated by opaque training processes and the often undisclosed inclusion of supervised data in contemporary Large Language Models (LLMs). This issue skews benchmark effectiveness and fosters potentially unfair comparisons, impeding the field's healthy development. To address this, we introduce a detection pipeline utilizing Perplexity and N-gram accuracy—two simple and scalable metrics that gauge a model's prediction precision on benchmark—to identify potential data leakages. By analyzing 31 LLMs under the context of mathematical reasoning, we reveal substantial instances of training even test set misuse, resulting in potentially unfair comparisons. These findings prompt us to offer several recommendations regarding model documentation, benchmark setup, and future evaluations. Notably, we propose the ``\emph{Benchmark Transparency Card}'' (Tab.~\ref{tab:benchmark-transparency-card}) to encourage clear documentation of benchmark utilization, promoting transparency and healthy developments of LLMs. we have made our leaderboard, pipeline implementation, and model predictions publicly available, fostering future research.

\textbf{Code}: \url{https://github.com/GAIR-NLP/benbench}

\textbf{Homepage}: \url{https://gair-nlp.github.io/benbench}

\textbf{Case Study Demo}: \url{https://huggingface.co/spaces/GAIR/benbench}

\end{abstract}

\section{Introduction}

The rapid development of large language models (LLMs) has resulted in a significant lag in the development of evaluation methods/protocols~\citep{10.1145/3641289-llm-evaluation-survey}. Coupled with the opacity of LLMs training, this can lead to a situation where it becomes challenging for individuals to form an objective assessment of evaluation results~\citep{DBLP:journals/corr/abs-2310-12941-transparency-index}. This overestimates the effectiveness of benchmarks, overlooks potential unfair comparison factors, and ultimately leads to missing scientifically meaningful directions, wasting societal resources. Especially, many models have explicitly involved supervised data in the pre-training phase, such as GLM-130B~\citep{DBLP:conf/iclr/ZengLDWL0YXZXTM23-glm-30b}, Qwen~\citep{qwen}, Nemotron-4 15B~\citep{DBLP:journals/corr/abs-2402-16819-Nemotron-4-15B}, InternLM-2~\citep{cai2024internlm2}, MiniCPM~\citep{Hu2024MiniCPM}, and etc. This context sets the stage for discussing the critical issue of benchmark data leakage. As reliance on these benchmarks grows, so does the risk that they may inadvertently be incorporated into the training data of LLMs,  thereby undermining evaluation integrity and complicating true capability assessments.

In exploring this issue, selecting an appropriate testbed is crucial. The ideal testbed should exhibit specific characteristics: (1) it should include both training and test sets, allowing for controlled comparisons; (2) improving performance on this benchmark should be inherently challenging, with limited effective datasets available. This scarcity increases the temptation for developers to use benchmark data to enhance performance; (3) it should also be of widespread interest, ensuring it is a standard metric for evaluating popular models (such as GPT-4~\citep{DBLP:journals/corr/abs-2303-08774-gpt-4}, Claude-3~\citep{claude-3}, etc.). Given these criteria, the mathematical reasoning  benchmark datasets GSM8K~\citep{DBLP:journals/corr/abs-2110-14168-gsm8k} and MATH~\citep{DBLP:conf/nips/HendrycksBKABTS21-math}\footnote{More benchmark datasets will be introduced gradually.} emerge as fitting choices for our test bed, which allow us to study data leakage in depth but also offer a relevant and challenging environment. Our primary aim with these datasets is to unearth potential benchmark leakage, enhancing transparency in language model development.

Given that training data and model details are often opaque, and leakage detection is influenced by various factors such as mode size and training strategies, detecting benchmark leakage is not a trivial task. In this work, we are not pursuing technical contributions in system development; instead, we are attempting to encourage the healthy development of this field, particularly through the lens of mathematical reasoning tasks, in the following aspects: \textbf{(1) summaries of various pre-training behaviors and challenges for detecting benchmark leakage} (cf. \S~\ref{sec:background}): Data leakage can occur in various scenarios and its detection is influenced by multiple factors such as unreliable assumptions, model size, training strategies, unknown training data, and even inaccessible model weights. \textbf{(2) proposal of a detection pipeline for estimating pre-training behaviors} (cf. \S~\ref{sec:detection_methodology}): We introduce a simple, computationally efficient, and scalable pipeline that leverages two fundamental yet insightful atomic metrics: \textit{Perplexity} and \textit{N-gram Accuracy}. These metrics effectively encapsulate the essence of language modeling, capturing its nuances from continuous and discrete perspectives, respectively. By paraphrasing benchmarks to create varied reference versions, we can detect discrepancies in models' atomic metrics, thereby identifying potential data leakage. This pipeline's validity is supported by thorough meta-experiments (cf. \S~\ref{sec:meta_experiment}). \textbf{(3) leakage analysis of existing models} (\S~\ref{sec:evaluation}): We extend our investigation to analyze existing models (i.e., 31 open-source LLMs), revealing that, in addition to previously identified leaks, many (i.e., approximately half of them), including well-known language models, may have inadvertently leveraged training data to boost their performance on mathematical reasoning tasks, leading to unfair advantages. Moreover, our metric even enables instance-level detection, revealing the possibility of test set leaks in many models (cf. \S~\ref{sec:instance-level-dection}). For example, we found that Qwen-1.8B can accurately predict all 5-grams in 223 examples from the GSM8K training set and 67 from the MATH training set, with an additional 25 correct predictions even in the MATH test set. \textbf{(4) recommendation for model documentation, benchmark setup and future evaluations} (cf. \S~\ref{sec: recommendation}): Based on these findings, we offer suggestions encompassing model documentation, benchmark construction, public access to benchmarks, and evaluation from multiple perspectives. We particularly emphasize the aspect of model documentation; we recommend that models should be accompanied by a document at release, which registers whether benchmark data was used for specific performance enhancement and whether any data augmentation was conducted. To this end, we introduce the \emph{Benchmark Transparency Card} (cf. \S~\ref{sec:benchmark-transparency-card} and Table~\ref{tab:benchmark-transparency-card}) to facilitate this process, hoping that it will be widely adopted to promote transparency and healthy development of LLMs.

These revelations underscore the urgency for a paradigm shift in how we approach the development and evaluation of language models. By pinpointing potential data leakage, our work champions greater transparency and fairness in model development, steering the community towards more ethical and effective research methodologies.

\section{Preliminaries}
\label{sec:background}

We summarize potential (pre-)training behaviors, define benchmark leakage, and highlight related challenges.

\subsection{Typical Training Behaviors}

\noindent\textbf{Training without seeing benchmark data} means that any benchmark data (including training and test splits) is not included in any training set of the model. This represents the most ideal scenario, where the model's performance on benchmarks stems from its generalized capabilities rather than overfitting to the benchmarks. The results on the benchmarks are genuine and reliable, aligning with the expectations of all stakeholders, including users and investors.

\noindent\textbf{Training with benchmark training data} means that the training split of benchmark data is (fully or partially) included in the training set of the model. This scenario could occur in situations where the training set of a benchmark is inadvertently mixed in during the training data collection process; it could also happen in efforts to specifically enhance certain capabilities of a model. For instance, GPT-4's training involved the integration of training datasets from GSM8K~\citep{DBLP:journals/corr/abs-2110-14168-gsm8k} and MATH~\citep{DBLP:conf/nips/HendrycksBKABTS21-math} to boost  its mathematical reasoning skills, as evidenced by their technical report~\citep{DBLP:journals/corr/abs-2303-08774-gpt-4}. Opaque training data and details can lead to unfair comparisons between models, especially for those that have not been exposed to the benchmark training splits. This opacity can also result in overly optimistic estimates of a model's generalization capabilities.

\noindent\textbf{Training with benchmark test data} means that the test split of the benchmark is fully or partially included in the training set of the model. Even though the model demonstrates satisfactory performance on benchmarks, this does not necessarily imply strong generalization capabilities. This situation could render the benchmarks ineffective, thereby misleading stakeholders such as users and downstream developers—a scenario we definitely want to avoid.

\vspace{-8pt}

\subsection{Definition of Benchmark Leakage}

Let $\mathcal{D}_{pretrain}$ be the dataset (corpus) used for pre-training a language model $\mathcal{M}$. For example $(x, y)$ from a benchmark $\mathcal{D}$ (regardless of whether it's from the training split $\mathcal{D}_{train}$ or the test split $\mathcal{D}_{test}$), if the example is found within $\mathcal{D}_{pretrain}$, it can be concluded that there has been a leak in the benchmark. Meanwhile, there are various ways in which a sample leak can occur, including (1) \emph{Input Leak}, where the input of the sample appears in $\mathcal{D}_{pretrain}$ and then the model is exposed to the question or prompt of the benchmark without corresponding answers; (2) \emph{Input-Output Leak}, where both the input and output of the sample are found in $\mathcal{D}_{pretrain}$.  This can confer an unfair advantage to models trained on such data.

\vspace{-8pt}

\subsection{Challenges for Detecting Benchmark Leakage}

Suppose that a benchmark has two splits: training and test. If one would like to detect whether the two splits are leaked, there are several challenges as follows: 

\textbf{(1) Cannot guarantee that the test data is leakage-free}: The most intuitive approach would be to compare certain metrics, such as perplexity, between the train and test splits. If a model has been exposed to the benchmark's data during (pre-)training, it will naturally exhibit a lower perplexity score on this data. Specifically, if the model has seen the training set but not the test set, its perplexity score on the training set will be lower than on the test set. Conversely, if the model has not seen either the training or test sets, its perplexity scores on both should be similar. However, identifying a model that has seen both the training and test sets poses a significant challenge because its perplexity scores may resemble those of a model that has seen neither, complicating the detection of data leakage. 

\textbf{(2) Difficult to determine the threshold score for leakage due to multiple influencing factors}: Even if a model has only been exposed to the training set, determining the extent of training set leakage by observing the difference in perplexity scores between the two sets is challenging. This difficulty arises because the threshold for detecting leakage can vary widely, influenced by factors such as model size, training strategies, and the distribution of the (pre-)training data (including the extent of data leakage), making it hard to establish a universal threshold. Additionally, retraining a model for each one under examination to simulate data leakage is impractical due to the enormous costs involved and the unknown distribution of training data. Moreover, what presents an even greater challenge is the fact that if a model is trained on a training set (as evidenced in \S~\ref{sec:meta_experiment}), it will exhibit a level of generalization on the test set, which makes it challenging to discern whether a model has been trained on both the training and test sets simultaneously.  

\textbf{(3) Unknown utilization of benchmarks}: There is limited knowing of how benchmark datasets are used during pre-training phase. The benchmark data might have been enhanced through various augmentation and data synthesis techniques, including paraphrasing and reformatting, among others~\citep{yu2024metamath,yang2023rethinking, fan2024reformatted}. Additionally, benchmarks might be employed for hyperparameter tuning during pre-training. 

\textbf{(4) Inaccessible model weights}: Many of today's powerful LLMs are closed-source, with access provided only through APIs. This typically results in inaccessible logit scores, posing significant challenges for calculating perplexity-related metrics.  Overall, these factors together significantly complicate the detection of benchmark leakage.

\section{Detection Methodology}
\label{sec:detection_methodology}

\subsection{Atomic Detection Metrics}

\noindent\textbf{Perplexity} \quad As one of the most common metrics for evaluation language model, it quantifies how well a language model predicts a sample of text~\citep{jelinek1977perplexity}. In essence, perplexity measures the uncertainty of a language model in predicting the next token (for example, a word or character) in a sequence. A lower perplexity score indicates that the language model is more confident in its predictions. Formally, it is defined as the exponentiated average negative log-likelihood of a sequence, expressed as:
\begin{equation}
    \text{PPL}(\boldsymbol{X}) = \exp (- \frac{1}{t} \sum_{i=0}^t\log p_\theta(x_i|x_{<i})),
\end{equation}
where $\boldsymbol{X} = [x_0, x_1, \dots, x_t]$ denotes a tokenized sequence. In the context of data leakage detection, we concatenate the question and answer part of a sample with a specific marker ``\texttt{ Answer: }'' and only calculate perplexity on the answer part of the combined text as the loss on the question part may not be considered during training.

\begin{figure}[t]
% \begin{wrapfigure}{r}{10cm}
\centering
\includegraphics[width=0.880\textwidth]{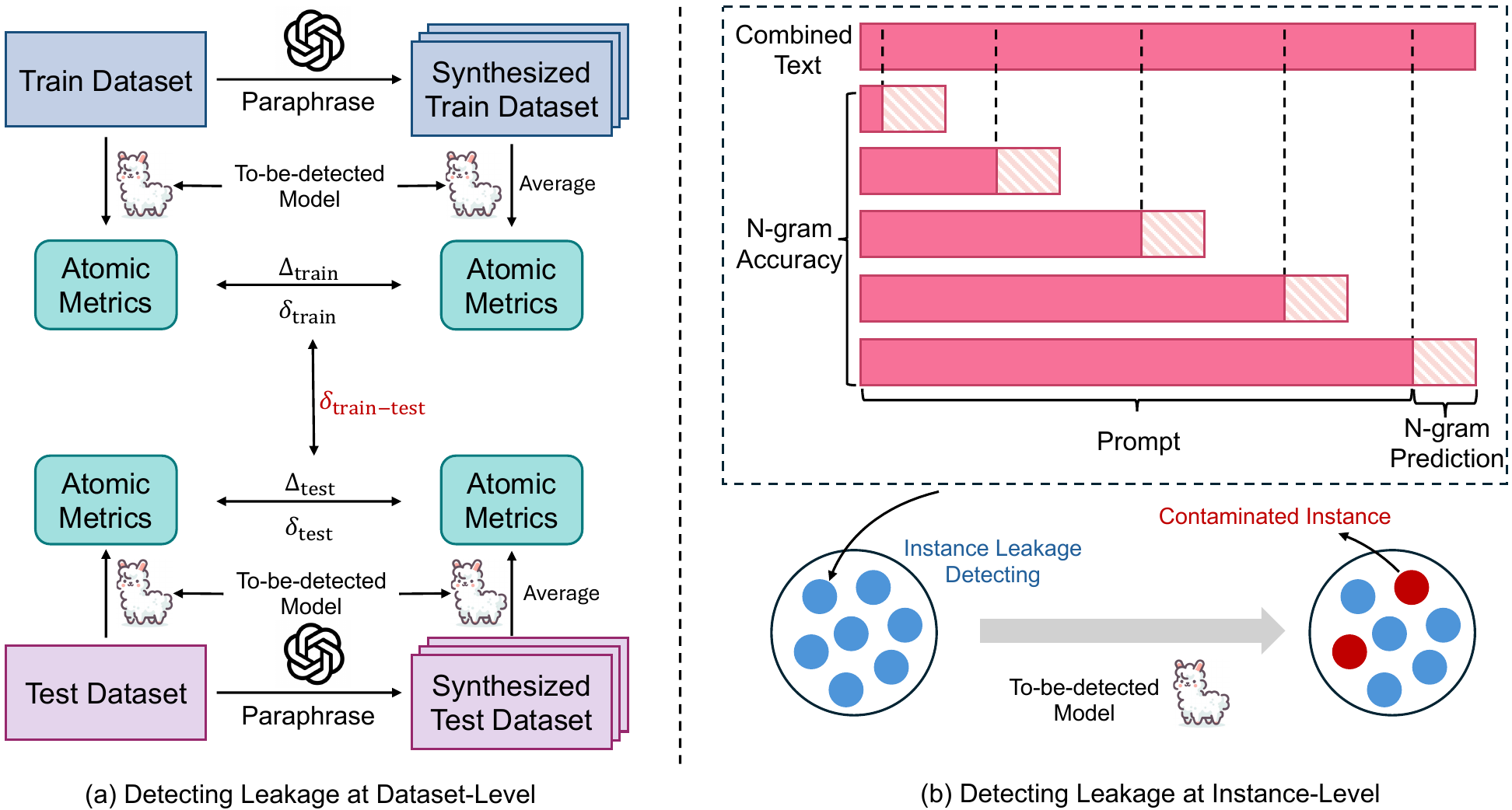}
\caption{An overview of our detecting approach.}
\label{fig:overview}
% \vspace{-12pt}
\end{figure}

\noindent\textbf{N-gram Accuracy} \quad We design another metric, called N-gram Accuracy, to help the fine-grained detection (cf. Figure~\ref{fig:overview} (b)). First, we combine the question and answer part with a single space for each sample, resulting in the combined text $X$. Second, we uniformly sample $K$ (i.e., 5) starting point among the interval from 2 to $|X|$.\footnote{The reason for starting from 2 is to prevent the model from generating without any conditions.} Then, the combined text from the beginning to the starting point is used as the prompt, with the subsequent $n$-gram serving as the target for prediction. 
If the majority of n-grams in a sample are correctly predicted, we can suspect that the model has already encountered this particular sample during training. 
As a result, this metric serves as a significant feature by aiding in the identification of potential instance-level data leakage (cf. \S~\ref{sec:instance-level-dection}). Formally, the n-gram accuracy on a dataset given a model can be expressed as:
\begin{equation}
    \text{N-gram Accuracy}(\boldsymbol{X}) = \frac{1}{S K} \sum_{i=0}^S  \sum_{j=0}^K I(X_{\text{start}_j:\text{start}_j+n}, \hat{X}_{\text{start}_j:\text{start}_j+n}),
\end{equation}
where the dataset size is $S$, $start_j$ denotes the index that $j$-th starting point corresponds to, and $X_{start_j:start_j+n}$ denotes the golden n-gram to be predicted, while $\hat{X}_{start_j:start_j+n}$ stands for the generated n-gram from a model $\mathcal{M}$ given the prompt. The indicator function $I$ applies an exact-match approach by default for precise measurement. Additionally, we consider using ROUGE-L~\citep{lin-2004-rouge} and edit distance similarity (cf. \S~\ref{appendix-sec:edit_distance}) to loosely measure the predicted n-grams, making the metric more robust to certain situations, such as when benchmark data has undergone some form of augmentation (paraphrasing~\citep{yang2023rethinking}, reformatting~\citep{fan2024reformatted}, etc.) (See \S~\ref{sec:instance-level-dection} for discussions). Note that all models 
adopt greedy decoding by default.

\noindent\textbf{A Unified View} \quad Both Perplexity and N-gram Accuracy metrics provide unique insights into a language model's performance, primarily assessing precision in next-token prediction. Perplexity, a continuous measure, evaluates the average likelihood of sequences, indicating model uncertainty. Conversely, N-gram Accuracy is a discrete metric focusing on the model's ability to replicate exact subsequences (n-grams) from training data. As Perplexity decreases—implying higher average log probabilities—the model's probability estimates for correct tokens increase, $p_{\mathcal{M}}(x_i|x_{<i})$. This enhancement in token-wise probabilities boosts the likelihood of precise n-gram predictions:
\begin{equation}
p(\text{n-gram}| \text{prompt}) = \prod_{k=0}^{n-1} p_{\mathcal{M}}(x_{start_j+k} | x_{<start_j+k}).
\end{equation}
Leveraging both Perplexity's subtle probabilistic nuances and N-gram Accuracy's exact match precision provides a comprehensive tool for evaluating model behavior under various conditions. This dual approach not only enhances understanding of model capabilities but also aids in detecting potential training data leakage and other model vulnerabilities. Furthermore, it is important to note that even when working with commercial closed-source models, where access to internal model outputs like logits is not possible, N-gram Accuracy can still be effectively used to analyze model behavior.

\subsection{Reference Benchmark Synthesis}

Given that we cannot guarantee the test set has remained contaminated, solely comparing the metric differences between training and test sets is unreliable. Ideally, we should refer to data that the model has definitively never encountered. A common approach is to use the most recently released examination questions as a benchmark~\citep{testing_language_models_on_a_held_out_high_school_national_finals_exam}. However, ensuring these questions are distributed similarly to the observed benchmark dataset (such as in terms of difficulty) is challenging, and finding suitable new questions for all benchmarks simultaneously is hardly feasible. Therefore, we propose the use of data synthesis, meaning that new reference benchmark is generated on-demand for each benchmark. This approach is designed to maintain the original benchmark's format and reasoning difficulty (such as numerical reasoning in math) by merely paraphrasing the surface text without altering its structure. This method acknowledges that synthesis may introduce biases in terms of phraseology; however, it aims to provide a viable alternative when fresh, untouched data is unavailable. Specifically, we utilize ChatGPT (\texttt{gpt-3.5-turbo-0125}) to create synthesized reference benchmarks for each existing benchmark. To minimize potential bias, we generate three distinct versions for each benchmark through generation sampling. The temperature is set to $0.7$ and top\_p to $0.9$ during sampling. We provide the full prompts for data synthesis in Table~\ref{tab:paraphrase_prompt_gsm8k} and Table~\ref{tab:paraphrase_prompt_math} and synthesized examples in Table~\ref{tab:paraphrase_example_GSM8K} and Table~\ref{tab:paraphrase_example_math}.

% \end{wrapfigure}

\subsection{Detection Pipeline}

Having introduced two atomic metrics and the synthesis of reference benchmarks, we will now orderly introduce our detection pipeline, as shown in Figure~\ref{fig:overview}.

\noindent\textbf{Step\#1 Preparation} \quad We synthesize three benchmark datasets based on the original benchmark, denoted as $\mathcal{D}_{\text{ref}_i}$ and select an atomic metric (e.g., N-gram Accuracy). Let  $\mathcal{M}$ be the model to be evaluated.

\noindent\textbf{Step\#2 Calculate the Atomic Metric on Benchmarks} \quad We calculate the atomic metric on both the original benchmark and the synthesized benchmark datasets, respectively. To mitigate randomness introduced during synthesis, we average the metrics across the three synthesized datasets. For all splits of the benchmark, we measure the decrement in the atomic metric, denoted by $\Delta$, which is the difference between the metric scores of the original benchmark ($M_{\text{ori}}$) and the synthesized benchmark ($M_{\text{ref}}$). The decrement $\Delta$ indicates the model's familiarity or memorization of the original benchmark relative to the synthesized set, demonstrating its capability to differentiate between potentially trained data and new, unseen data. The direction of subtraction is dependent on the nature of the atomic metric; for instance, if using perplexity, the calculation would be reversed ($M_{\text{ref}} - M_{\text{ori}}$). Given that the decrement $\Delta$ might be significantly influenced by the model's size and capability, leading to large numerical discrepancies, we normalize this metric by dividing it with the original metric score, $M_{\text{ori}}$, to standardize it. This results in the percentage decrease in the atomic metric $\delta$, expressed as:
\begin{equation}
    \Delta = M_\text{ori} - M_\text{ref}, \quad 
    \delta =  \frac{\Delta}{M_\text{ori}} \times 100\%. 
\end{equation}
This normalization allows for meaningful comparisons across different models. For example, consider a model $\mathcal{M}$ that achieves 38.47\% and 21.52\% in 5-gram accuracy on the training set of the original and synthesized benchmarks, respectively. This results in a difference $\Delta_{\text{train}}$ of 16.95, and a relative decrease $\delta_{\text{train}}$ of 44.06\%.

\noindent\textbf{Step\#3 Comparison} \quad  We analyze $\delta_\text{train}$ and $\delta_\text{test}$ and compute the disparity $\delta_{\text{train-test}} = \delta_{\text{train}} - \delta_{\text{test}}$. This disparity, derived by subtracting $\delta_{\text{test}}$ from $\delta_{\text{train}}$, offsets biases introduced by data synthesis and depicts the model's relative familiarity and memorization of the training set compared to the test set. If this disparity approaches zero, it indicates a consistent relative decline across both training and test datasets, suggesting that the degree of leakage is equivalent in both splits—either no leakage or simultaneous leakage.\footnote{In cases where both the training and testing sets leak simultaneously, our pipeline may not effectively detect it. Fortunately, our n-gram accuracy metric can greatly mitigate this issue (cf. \S~\ref{sec:instance-level-dection}).} Conversely, a significant $\delta_{\text{train-test}}$ suggests the model is disproportionately familiar with the training dataset compared to the test set, indicating potential leakage in the training set. However, it should be noted that the possibility of leakage in the test set still cannot be entirely ruled out. A notably negative $\delta_{\text{train-test}}$ points to potential leakage in the test dataset. For instance, a model that records a $\delta_{\text{train-test}}$ of 35.54\% implies a higher likelihood of data leakage in the training set compared to another model with a $\delta_{\text{train-test}}$ of 0.96\%.

\section{Meta-Experiment: the Reliability of the Detecting Pipeline}
\label{sec:meta_experiment}

\noindent\textbf{Setup} \quad Considering the significant cost associated with integrating benchmarks into large-scale pre-training simulations, as well as the performance dependency on variables such as model size and data volume~\citep{DBLP:journals/corr/abs-2401-06059-investigating-data-contamination}, we have opted for a more straightforward simulation way to estimate the upper bound, entailing direct training of the language model using benchmark data, implemented in two distinct methods: (1) \emph{Pre-training}: The model is pre-trained on the benchmark data where the full loss is utilized; (2) \emph{Supervised Fine-tuning} (SFT): The model is fine-tuned on the benchmark data, but the loss is calculated solely on the solution portion.\footnote{It simulates scenarios where models may have potentially undergone supervised fine-tuning or employed the prefix language modeling pre-training objective ~\citep{10.5555/3455716.3455856-t5} on the benchmark.} To validate the effectiveness of our detection pipeline across different training strategies, we prepare two sets of data from each benchmark: (i) A ``seen'' set, comprising 1,000 entries used for model training; (ii) An ``unseen'' set, consisting of another 1,000 entries that the model has not been exposed to.  Then both sets undergo paraphrasing synthesis to generate three distinct reference sets for the ``seen'' and ``unseen'' sets, respectively. We employ the \texttt{Mistral-7B-v0.1}~\citep{jiang2023mistral}  model for these experiments. Further training details can be found in  Appendix~\ref{appendix-sec:meta-experiments}. By applying our detection pipeline to the resulting models under different training strategies, we can controllably validate the effectiveness of our approach.

\begin{figure}[h]
\centering
\begin{tikzpicture}
  % Define the style for the secondary axis
  \pgfplotsset{
    secondary axis style/.style={
      axis y line*=right,
      axis x line=none,
      ylabel near ticks, ylabel shift = -5 pt
    },
  }

  % First subplot
  \begin{axis}[
      title={5-gram Acc. on GSM8K},
      width=4.3cm,
      height=5cm,
      ymin=0, ymax=12,
      ybar,
      bar width=10pt,
      symbolic x coords={Mistral, + SFT, + Pretrain},
      xtick=data,
      grid style=dashed,
      axis y line*=left,
      ylabel={Decrement $\Delta$},
      ylabel near ticks,
      ylabel shift = -4 pt,
      yticklabel shift = -2 pt,
      enlarge x limits=0.25,
      xtick pos=bottom, % Only show xticks at the bottom
      ytick pos=left,   % Only show yticks at the left
      at={(0,0)}, % Position of the first subplot
      ylabel style = {font=\tiny},
      xticklabel style={font=\tiny},
      yticklabel style={font=\tiny},
      title style={font=\tiny, yshift=-8pt},
  ]
    \addplot+[ybar, fill=lightgreen, fill opacity = 0.8, draw=none] coordinates {(Mistral, 1.29) (+ SFT, 8.62)  (+ Pretrain, 10.33)};
    \addplot+[ybar, fill=darkgreen, fill opacity = 0.8, draw=none] coordinates {(Mistral, 0.45)  (+ SFT, 3.84) (+ Pretrain, 5.21)};
  \end{axis}

  \begin{axis}[
      width=4.3cm,
      height=5cm,
      ymin=0, ymax=24, % Adjust the max value based on your line plot data
      enlarge x limits=0.25,
      symbolic x coords={Mistral, + SFT, + Pretrain},
      secondary axis style,
      at={(0,0)}, % Position of the first subplot
      ylabel style = {font=\tiny},
      yticklabel shift = -2 pt,
      xticklabel style={font=\tiny},
      yticklabel style={font=\tiny},
      title style={font=\tiny, yshift=-8pt},
      grid style=dashed,
  ]
    \addplot[sharp plot, lightgreen, opacity=0.5, line width=0.7pt, mark=x, mark options={solid, scale=1.0}] coordinates { (Mistral, 7.56) (+ SFT, 17.88) (+ Pretrain, 20.41)};
    \addplot[sharp plot, darkgreen, opacity=0.5, line width=0.7pt, mark=x, mark options={solid, scale=1.0}] coordinates {(Mistral, 2.71) (+ SFT, 9.13) (+ Pretrain, 11.97)};
    \addplot[sharp plot, purple, opacity=0.8, line width=0.7pt, mark=o, mark options={solid, scale=0.8}] coordinates {(Mistral, 4.85) (+ SFT, 8.75) (+ Pretrain, 8.44)};
  \end{axis}

  % second subplot
  \begin{axis}[
      title={5-gram Acc. on MATH},
      width=4.3cm,
      height=5cm,
      ymin=0, ymax=12,
      ybar,
      bar width=10pt,
      symbolic x coords={Mistral, + SFT, + Pretrain},
      xtick=data,
      axis y line*=left,
      enlarge x limits=0.25,
      at={(3.3cm,0)}, % Position of the first subplot
      yticklabel shift = -2 pt,
      xticklabel style={font=\tiny},
      yticklabel style={font=\tiny},
      title style={font=\tiny, yshift=-8pt},
      xtick pos=bottom, % Only show xticks at the bottom
      ytick pos=left,   % Only show yticks at the left
  ]
   \addplot+[ybar, fill=lightgreen, draw=none, fill opacity = 0.8] coordinates {(Mistral, 2.55) (+ SFT, 8.87)  (+ Pretrain, 9.83)};
    \addplot+[ybar, fill=darkgreen, draw=none, fill opacity = 0.8] coordinates {(Mistral, 2.45)  (+ SFT, 2.73) (+ Pretrain, 3.9)};
  \end{axis}
  
  \begin{axis}[
      width=4.3cm,
      height=5cm,
      ymin=-1, ymax=24, % Adjust the max value based on your line plot data
      enlarge x limits=0.25,
      symbolic x coords={Mistral, + SFT, + Pretrain},
      secondary axis style,
      yticklabel shift = -2 pt,
      at={(3.3cm,0)}, 
      xticklabel style={font=\tiny},
      yticklabel style={font=\tiny},
      title style={font=\tiny},
  ]
     \addplot[sharp plot, lightgreen, opacity=0.5, line width=0.7pt, mark=x, mark options={solid, scale=1.0}] coordinates { (Mistral, 7.67) (+ SFT, 18.09) (+ Pretrain, 19.17)};
    \addplot[sharp plot, darkgreen, opacity=0.5, line width=0.7pt, mark=x, mark options={solid, scale=1.0}] coordinates {(Mistral, 7.74) (+ SFT, 8.02) (+ Pretrain, 10.94)  };
    \addplot[sharp plot, purple, opacity=0.8, line width=0.7pt, mark=o, mark options={solid, scale=0.8}] coordinates {(Mistral, -0.07) (+ SFT, 10.07) (+ Pretrain, 8.23)};
  \end{axis}

    %  3 plot
  \begin{axis}[
      title={PPL on GSM8K},
      width=4.3cm,
      height=5cm,
      ymin=-0.1, ymax=0.8,
      ybar,
      bar width=10pt,
      symbolic x coords={Mistral, + SFT, + Pretrain},
      xtick=data,
      axis y line*=left,
      yticklabel shift = -2 pt,
      enlarge x limits=0.25,
      at={(6.6cm,0)}, % Position of the first subplot
      xticklabel style={font=\tiny},
      yticklabel style={font=\tiny},
      title style={font=\tiny, yshift=-7.5pt},
      xtick pos=bottom, % Only show xticks at the bottom
      ytick pos=left,   % Only show yticks at the left
  ]
    \addplot+[ybar, fill=lightgreen, draw=none, fill opacity = 0.8] coordinates {(Mistral, -0.01) (+ SFT, 0.54)  (+ Pretrain, 0.46)};
    \addplot+[ybar, fill=darkgreen, draw=none, fill opacity = 0.8] coordinates {(Mistral, 0.02)  (+ SFT, 0.35) (+ Pretrain, 0.3)};
  \end{axis}
  
  \begin{axis}[
      width=4.3cm,
      height=5cm,
      ymin=-5, ymax=44, % Adjust the max value based on your line plot data
      enlarge x limits=0.25,
      symbolic x coords={Mistral, + SFT, + Pretrain},
      secondary axis style,
      yticklabel shift = -2 pt,
      at={(6.6cm,0)}, % Position of the first subplot
      xticklabel style={font=\tiny},
      yticklabel style={font=\tiny},
      title style={font=\tiny}
  ]
     \addplot[sharp plot, lightgreen, opacity=0.5, line width=0.7pt, mark=x, mark options={solid, scale=1.0}] coordinates { (Mistral, -0.31) (+ SFT, 41.22) (+ Pretrain, 34.33)};
    \addplot[sharp plot, darkgreen, opacity=0.5, line width=0.7pt, mark=x, mark options={solid, scale=1.0}] coordinates {(Mistral, 0.63) (+ SFT, 21.88) (+ Pretrain, 18.99)  };
    \addplot[sharp plot, purple, opacity=0.8, line width=0.7pt, mark=o, mark options={solid, scale=0.8}] coordinates {(Mistral, -0.94) (+ SFT, 19.34) (+ Pretrain, 15.34)};
    
  \end{axis}

   %  4 plot
  \begin{axis}[
      title={PPL on MATH},
      width=4.3cm,
      height=5cm,
      ymin=0, ymax=0.8,
      ybar,
      bar width=10pt,
      symbolic x coords={Mistral, + SFT, + Pretrain},
      xtick=data,
      axis y line*=left,
      yticklabel shift = -2 pt,
      enlarge x limits=0.25,
      at={(10cm,0)}, % Position of the first subplot
      xticklabel style={font=\tiny},
      yticklabel style={font=\tiny},
      title style={font=\tiny, yshift=-7.5pt},
      legend style={at={(1.05,1)}, anchor=south west}, 
      xtick pos=bottom, % Only show xticks at the bottom
      ytick pos=left,   % Only show yticks at the left
  ]
  \addplot+[ybar, fill=lightgreen,   draw=none, fill opacity = 0.8] coordinates {(Mistral, 0.3) (+ SFT, 0.58)  (+ Pretrain, 0.55)};
    % \addlegendentry{Seen} 
    \addplot+[ybar, fill=darkgreen, draw=none, fill opacity = 0.8] coordinates {(Mistral, 0.29)  (+ SFT, 0.34) (+ Pretrain, 0.35)};
    % \addlegendentry{Unseen} 
  \end{axis}
  
  \begin{axis}[
      width=4.3cm,
      height=5cm,
      ymin=0, ymax=44, % Adjust the max value based on your line plot data
      enlarge x limits=0.25,
      symbolic x coords={Mistral, + SFT, + Pretrain},
      secondary axis style,
      ylabel={Percentage decrease $\delta$},
      ylabel near ticks,
      ylabel shift = -4 pt,
      yticklabel shift = -2 pt,
      at={(10.cm,0)}, % Position of the first subplot
      xticklabel style={font=\tiny},
      yticklabel style={font=\tiny},
      title style={font=\tiny, yshift=-8pt},
      ylabel style = {font=\tiny},
  ]
  \addplot[sharp plot, lightgreen,  opacity=0.5, line width=0.7pt, mark=x, mark options={solid, scale=1.0}] coordinates { (Mistral, 12.45) (+ SFT, 37.91) (+ Pretrain, 35.26)};
    \addplot[sharp plot, darkgreen,  opacity=0.5, line width=0.7pt, mark=x, mark options={solid, scale=1.0}] coordinates {(Mistral, 11.65) (+ SFT, 14.66) (+ Pretrain, 15.22)};
     \addplot[sharp plot, purple, opacity=0.8, line width=0.7pt, mark=o, mark options={solid, scale=0.8}] coordinates {(Mistral, 0.8) (+ SFT, 23.25) (+ Pretrain, 20.04)};
  \end{axis}

    \begin{axis}[
        hide axis, 
        xmin=0, xmax=1, ymin=0, ymax=0.1, 
        legend columns=-1, 
        legend style={
            draw=none, 
            at={(1,-0.1)}, 
            anchor=north, 
            font=\footnotesize, 
        }
    ]
    \addlegendimage{ybar,ybar legend,fill=lightgreen,draw=none}
    \addlegendentry{$\Delta_\text{seen}$}
    \addlegendimage{ybar,ybar legend,fill=darkgreen,draw=none}
    \addlegendentry{$\Delta_\text{unseen}$}
    \addlegendimage{sharp plot,lightgreen, mark=x, mark options={solid, scale=1.0}}
    \addlegendentry{$\delta_\text{seen}$}
    \addlegendimage{sharp plot,darkgreen, mark=x, mark options={solid, scale=1.0}}
    \addlegendentry{$\delta_\text{unseen}$}
    \addlegendimage{sharp plot,purple, mark=o, mark options={solid, scale=1.0}}
    \addlegendentry{$\delta_\text{seen-unseen}$}
    \end{axis}
\end{tikzpicture}
    \vspace{-15pt}
 \caption{Meta experiment results on GSM8K and MATH datasets. The left y-axis indicates the decrement $\Delta$, and the right y-axis shows the percentage decrease $\delta$.}
\label{fig:meta_results}
\end{figure}
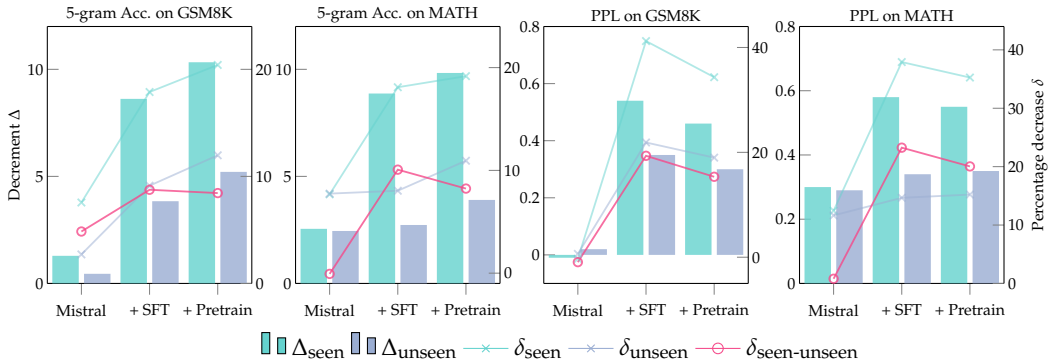

\noindent\textbf{Results} \quad As shown in Figure~\ref{fig:meta_results}, whether the model is trained on benchmark data via supervised fine-tuning or pre-training, it becomes more familiar with the data, leading to a greater difference in atomic metrics ($\Delta_\text{seen}$ and $\delta_\text{seen}$) between the original and the synthetic reference benchmarks. This also results in some generalization ability on unseen benchmark data, causing an increase in both $\Delta_\text{unseen}$ and $\delta_\text{unseen}$. However, if the model has not been trained on any benchmark data, these metrics will be low, and the difference between $\delta_\text{seen}$ and $\delta_\text{unseen}$ (\tikz[baseline={([yshift=-3pt]char.center)}]{ 
    \node[shape=circle,draw,purple,inner sep=1.5pt] (char) {}; 
    \draw[purple, line width=0.5pt] ([xshift=-7.5pt]char.west) -- ([xshift=7.5pt]char.east);
}) will essentially be zero. Meanwhile, 
as the calculation of n-gram accuracy spans the entire text of both question and solution, it yields in higher  $\Delta$ and $\delta$ scores for models trained through pre-training compared to those trained via supervised fine-tuning. In contrast, as perplexity is calculated only on the solution part, supervised fine-tuned models exhibit higher  $\Delta$ and $\delta$ scores than those pre-trained. Interestingly, supervised fine-tuning results in higher $\delta_\text{seen-unseen}$ scores compared to pre-training on benchmark data, which suggests that the model fits the training benchmark data better and improves the ease of detecting data leaks with supervised fine-tuning. Most importantly, the order of $\delta_\text{seen-unseen}$ scores is consistent across different atomic metrics and benchmark datasets. Overall, these results demonstrate the effectiveness of our pipeline.

\section{Evaluation in the Wild}
\label{sec:evaluation}

In this section, we employ our detection pipeline to existing language models.

\subsection{Setup}

\noindent\textbf{Evaluated LLMs}\quad We carried out an extensive evaluation of prominent large-scale language models, encompassing a diverse range including the GPT-4~\citep{DBLP:journals/corr/abs-2303-08774-gpt-4}, Claude-3~\citep{claude-3}, Grok-1~\citep{grok-1}, Qwen family~\citep{qwen}, Aquila2 family~\citep{aquila2}, InternLM family~\citep{2023internlm}, Baichuan family~\citep{baichuan2023baichuan2}, ChatGLM family~\citep{chatglm6b}, Yi family~\citep{ai2024yi}, LLaMA family~\citep{Touvron2023LLaMA,Touvron2023Llama2,Llama-3}, Gemma family~\citep{Mesnard2024Gemma}, Mistral-7B~\citep{jiang2023mistral}, Grok-1~\citet{grok-1}, Phi family~\citep{Li2023phi-1.5,Phi-2},  Orca-2-7B~\citep{DBLP:journals/corr/abs-2311-11045-orca-2}, DeepSeekMath~\citep{DBLP:journals/corr/abs-2402-03300-deepseek-math}. In total, we evaluate 31 models across a range of sizes.

\noindent\textbf{Benchmark Datasets}\quad We continue to utilize GSM8K and MATH for detecting data leaks. Specifically, for each dataset, we randomly select 3,000 samples from the training and test sets (or all if the number of samples is less than 3,000). We concatenate the question and solution with a space delimiter, thus forming a complete text sample. We calculate perplexity only on the solution part to avoid potential issues with the model not calculating loss on the question part. For n-gram accuracy, we compute across the combined text using n-gram values of 5 and 10.

\subsection{Main Results}
\label{sec:main-results}

\begin{figure}[h]
\centering
\begin{tikzpicture}
  % Define the style for the secondary axis
  \pgfplotsset{
    secondary axis style/.style={
      axis y line*=right,
      axis x line=none,
      ylabel near ticks, ylabel shift = -5 pt
    },
    }

  % First subplot
  \begin{axis}[
      width=14.2cm,
      height=5cm,
      ymin=-2, ymax=35,
      ybar,
      bar width=5pt,
      symbolic x coords={Qwen-7B, Qwen-14B, InternLM2-7B, InternLM2-20B, Aquila2-7B, Aquila2-34B,  Qwen-1.8B, Baichuan2-13B-Base, ChatGLM2-6B, Yi-34B, InternLM-20B, Phi-2, Yi-6B, Orca-2-7B, Phi-1.5, InternLM-7B, LLaMA2-7B, Aquila-7B, Baichuan-7B, ChatGLM3-6B,  LLaMA-7B, Mistral-7B-v0.1, Grok-1, Gemma-2B, Gemma-7B, Baichuan2-7B-Base, Llama-3-8B, InternLM2-20B-Base, Baichuan-13B-Base, DeepSeekMath-7B, InternLM2-7B-Base},
      xtick=data,
      grid style=dashed,
      axis y line*=left,
      ylabel={Decrement $\Delta$},
      ylabel near ticks,
      ylabel shift = -4 pt,
      yticklabel shift = -2 pt,
      enlarge x limits=0.05,
      xtick pos=bottom, % Only show xticks at the bottom
      ytick pos=left,   % Only show yticks at the left
      at={(0,0)}, % Position of the first subplot
      ylabel style = {font=\tiny},
      xticklabel style={ rotate=45, anchor=east, font=\tiny},
      yticklabel style={font=\tiny},
      title style={font=\tiny, yshift=-8pt},
  ]
    % seen
    \addplot+[ybar, bar shift=-3pt, fill=lightgreen, fill opacity = 0.8, draw=none] coordinates {
    (Qwen-7B, 16.95)
    (Qwen-14B, 24.32)
    (InternLM2-7B, 17.03)
    (InternLM2-20B, 18.27)
    (Aquila2-7B, 15.82)
    (Aquila2-34B, 20.18)
    (Qwen-1.8B, 32.22)
    (Baichuan2-13B-Base, 10.80)
    (ChatGLM2-6B, 6.29)
    (Yi-34B, 6.07)
    (InternLM-20B, 6.56)
    (Phi-2, 2.83)
    (Yi-6B, 4.30)
    (Orca-2-7B, 10.42)
    (Phi-1.5, 1.42)
    (InternLM-7B, 3.50)
    (LLaMA2-7B, 0.93)
    (Aquila-7B, 8.40)
    (Baichuan-7B, 0.87)
    (ChatGLM3-6B, 7.29)
    (LLaMA-7B, 0.78)
    (Mistral-7B-v0.1, 1.06)
    (Grok-1,0.74)
    (Gemma-2B, -0.10)
    (Gemma-7B, 0.57)
    (Baichuan2-7B-Base, 0.75)
    (Llama-3-8B, 0.24)
    (InternLM2-20B-Base, 0.80)
    (Baichuan-13B-Base, 0.57)
    (DeepSeekMath-7B, 0.94)
    (InternLM2-7B-Base, -0.19)
    };
    % unseen
    \addplot+[ybar, bar shift=2pt, fill=darkgreen, fill opacity = 0.8, draw=none] coordinates {
    (Qwen-7B, 1.90)
    (Qwen-14B, 3.93)
    (InternLM2-7B, 3.82)
    (InternLM2-20B, 4.43)
    (Aquila2-7B, 5.21)
    (Aquila2-34B, 6.93)
    (Qwen-1.8B, 11.49)
    (Baichuan2-13B-Base, 3.45)
    (ChatGLM2-6B, 2.25)
    (Yi-34B, 2.71)
    (InternLM-20B, 3.42)
    (Phi-2, 0.59)
    (Yi-6B, 2.18)
    (Orca-2-7B, 7.33)
    (Phi-1.5, 0.60)
    (InternLM-7B, 2.57)
    (LLaMA2-7B, 0.33)
    (Aquila-7B, 6.92)
    (Baichuan-7B, 0.65)
    (ChatGLM3-6B, 6.62)
    (LLaMA-7B, 0.62)
    (Mistral-7B-v0.1, 0.90)
    (Grok-1, 0.86)
    (Gemma-2B, -0.01)
    (Gemma-7B, 0.96)
    (Baichuan2-7B-Base, 1.08)
    (Llama-3-8B,0.53)
    (InternLM2-20B-Base, 1.15)
    (Baichuan-13B-Base, 0.95)
    (DeepSeekMath-7B, 1.46)
    (InternLM2-7B-Base, 0.20)
    };
  \end{axis}

  \begin{axis}[
      width=14.2cm,
      height=5cm,
      ymin=-5, ymax=60, % Adjust the max value based on your line plot data
      enlarge x limits=0.05,
      symbolic x coords={Qwen-7B, Qwen-14B, InternLM2-7B, InternLM2-20B, Aquila2-7B, Aquila2-34B,  Qwen-1.8B, Baichuan2-13B-Base, ChatGLM2-6B, Yi-34B, InternLM-20B, Phi-2, Yi-6B, Orca-2-7B, Phi-1.5, InternLM-7B, LLaMA2-7B, Aquila-7B, Baichuan-7B, ChatGLM3-6B,  LLaMA-7B, Mistral-7B-v0.1, Grok-1, Gemma-2B, Gemma-7B, Baichuan2-7B-Base, Llama-3-8B, InternLM2-20B-Base, Baichuan-13B-Base, DeepSeekMath-7B, InternLM2-7B-Base},
      secondary axis style,
      ylabel={Percentage decrease $\delta$},
      at={(0,0)}, % Position of the first subplot
      ylabel style = {font=\tiny},
      yticklabel shift = -2 pt,
      xticklabel style={font=\tiny},
      yticklabel style={font=\tiny},
      title style={font=\tiny, yshift=-8pt},
      grid style=dashed,
  ]
  % seen
    \addplot[sharp plot, lightgreen, opacity = 0.5, line width=0.7pt, mark=x, mark options={solid, scale=1.0}] coordinates {
    (Qwen-7B, 44.06)
    (Qwen-14B, 50.49)
    (InternLM2-7B, 48.99)
    (InternLM2-20B, 50.21)
    (Aquila2-7B, 55.14)
    (Aquila2-34B, 54.28)
    (Qwen-1.8B, 52.60)
    (Baichuan2-13B-Base, 31.17)
    (ChatGLM2-6B, 24.47)
    (Yi-34B, 24.60)
    (InternLM-20B, 28.26)
    (Phi-2, 14.10)
    (Yi-6B, 20.79)
    (Orca-2-7B, 29.75)
    (Phi-1.5, 8.34)
    (InternLM-7B, 18.95)
    (LLaMA2-7B, 6.03)
    (Aquila-7B, 23.68)
    (Baichuan-7B, 6.50)
    (ChatGLM3-6B, 22.73)
    (LLaMA-7B, 5.53)
    (Mistral-7B-v0.1, 6.21)
    (Grok-1, 4.54)
    (Gemma-2B, -0.63)
    (Gemma-7B, 3.04)
    (Baichuan2-7B-Base, 4.86)
    (Llama-3-8B,1.87)
    (InternLM2-20B-Base, 5.93)
    (Baichuan-13B-Base, 3.74)
    (DeepSeekMath-7B, 5.45)
    (InternLM2-7B-Base, -1.59)
    };
    % unseen
    \addplot[sharp plot, darkgreen, opacity = 0.5, line width=0.7pt, mark=x, mark options={solid, scale=1.0}] coordinates {
    (Qwen-7B, 8.52)
    (Qwen-14B, 15.33)
    (InternLM2-7B, 18.87)
    (InternLM2-20B, 21.10)
    (Aquila2-7B, 31.90)
    (Aquila2-34B, 31.16)
    (Qwen-1.8B, 29.73)
    (Baichuan2-13B-Base, 13.32)
    (ChatGLM2-6B, 10.82)
    (Yi-34B, 13.05)
    (InternLM-20B, 17.58)
    (Phi-2, 3.50)
    (Yi-6B, 11.74)
    (Orca-2-7B, 23.30)
    (Phi-1.5, 3.88)
    (InternLM-7B, 15.09)
    (LLaMA2-7B, 2.21)
    (Aquila-7B, 21.11)
    (Baichuan-7B, 4.84)
    (ChatGLM3-6B, 21.16)
    (LLaMA-7B, 4.46)
    (Mistral-7B-v0.1, 5.25)
    (Grok-1, 5.08)
    (Gemma-2B, -0.06)
    (Gemma-7B, 4.95)
    (Baichuan2-7B-Base, 6.92)
    (Llama-3-8B, 4.00)
    (InternLM2-20B-Base, 8.35)
    (Baichuan-13B-Base, 6.32)
    (DeepSeekMath-7B, 8.09)
    (InternLM2-7B-Base, 1.62)
    };
    %  seen- unseen
    \addplot[sharp plot, purple, opacity = 0.8, line width=0.7pt, mark=o, mark options={solid, scale=0.8}] coordinates {
    (Qwen-7B, 35.54)
    (Qwen-14B, 35.16)
    (InternLM2-7B, 30.12)
    (InternLM2-20B, 29.11)
    (Aquila2-7B, 23.24)
    (Aquila2-34B, 23.12)
    (Qwen-1.8B, 22.87)
    (Baichuan2-13B-Base, 17.85)
    (ChatGLM2-6B, 13.65)
    (Yi-34B, 11.55)
    (InternLM-20B, 10.68)
    (Phi-2, 10.60)
    (Yi-6B, 9.05)
    (Orca-2-7B, 6.45)
    (Phi-1.5, 4.46)
    (InternLM-7B, 3.86)
    (LLaMA2-7B, 3.82)
    (Aquila-7B, 2.57)
    (Baichuan-7B, 1.66)
    (ChatGLM3-6B, 1.57)
    (LLaMA-7B, 1.07)
    (Mistral-7B-v0.1, 0.96)
    (Grok-1,-0.54)
    (Gemma-2B, -0.57)
    (Gemma-7B, -1.91)
    (Baichuan2-7B-Base, -2.06)
    (Llama-3-8B, -2.13)
    (InternLM2-20B-Base, -2.42)
    (Baichuan-13B-Base, -2.58)
    (DeepSeekMath-7B, -2.64)
    (InternLM2-7B-Base, -3.21)
    };
  \end{axis}

    \begin{axis}[
        hide axis, 
        xmin=0, xmax=1, ymin=0, ymax=0.1, 
        legend columns=-1, 
        legend style={
            draw=none, 
            at={(1.38, 0.58)}, 
            anchor=north, 
            font=\tiny, 
        }
    ]
    \addlegendimage{ybar,ybar legend,fill=lightgreen,draw=none}
    \addlegendentry{$\Delta_\text{train}$}
    \addlegendimage{ybar,ybar legend,fill=darkgreen,draw=none}
    \addlegendentry{$\Delta_\text{test}$}
    \addlegendimage{sharp plot,lightgreen, mark=x, mark options={solid, scale=1.0}}
    \addlegendentry{$\delta_\text{train}$}
    \addlegendimage{sharp plot,darkgreen, mark=x, mark options={solid, scale=1.0}}
    \addlegendentry{$\delta_\text{test}$}
    \addlegendimage{sharp plot, purple, mark=o, mark options={solid, scale=0.8}}
    \addlegendentry{$\delta_\text{train-test}$}
    \end{axis}
\end{tikzpicture}
\vspace{-15pt}
\caption{ LLMs ordered by $\delta_{\text{train-test}}$ w.r.t 5-gram accuracy on GSM8K. The left y-axis indicates the decrement $\Delta$, and the right y-axis shows the percentage decrease $\delta$. Models positioned on the left are more likely to train on the training set compared to the test set, with this likelihood diminishing as one moves to the right.}
\label{fig:main_results(5gram-G)}
\end{figure}
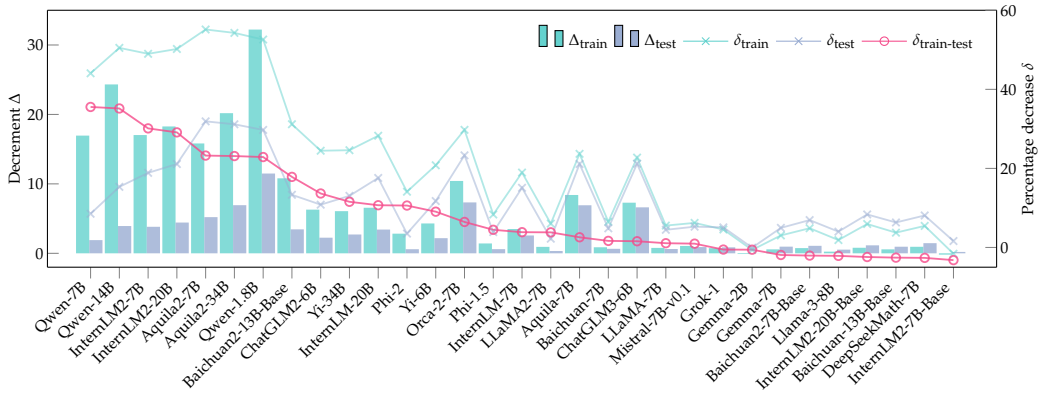
\begin{figure}[h]
\centering
\begin{tikzpicture}
  % Define the style for the secondary axis
  \pgfplotsset{
    secondary axis style/.style={
      axis y line*=right,
      axis x line=none,
      ylabel near ticks, ylabel shift = -5 pt
    },
    }

  % First subplot
  \begin{axis}[
      width=14.2cm,
      height=5cm,
      ymin=-0.6, ymax=3.5,
      ybar,
      bar width=5pt,
      symbolic x coords={Aquila2-7B, InternLM2-20B, InternLM2-7B, Aquila2-34B, ChatGLM2-6B, Qwen-14B, Qwen-7B, Orca-2-7B, InternLM-20B, Baichuan2-13B-Base, Qwen-1.8B, Yi-34B, Aquila-7B, ChatGLM3-6B, Yi-6B, InternLM-7B, Grok-1, LLaMA2-7B, Mistral-7B-v0.1,  Gemma-7B, InternLM2-7B-Base, LLaMA-7B, Llama-3-8B, Baichuan-13B-Base, DeepSeekMath-7B, Baichuan-7B, InternLM2-20B-Base, Baichuan2-7B-Base, Gemma-2B, Phi-1.5, Phi-2},
      xtick=data,
      grid style=dashed,
      axis y line*=left,
      ylabel={Decrement $\Delta$},
      ylabel near ticks,
      ylabel shift = -4 pt,
      yticklabel shift = -2 pt,
      enlarge x limits=0.05,
      xtick pos=bottom, % Only show xticks at the bottom
      ytick pos=left,   % Only show yticks at the left
      at={(0,0)}, % Position of the first subplot
      ylabel style = {font=\tiny},
      xticklabel style={ rotate=45, anchor=east, font=\tiny},
      yticklabel style={font=\tiny},
      title style={font=\tiny, yshift=-8pt},
  ]
    % seen
    \addplot+[ybar, bar shift=-3pt, fill=lightgreen, fill opacity = 0.8, draw=none] coordinates {
    (Aquila2-7B, 3.10)
    (InternLM2-20B, 1.49)
    (InternLM2-7B, 1.56)
    (Aquila2-34B, 1.31)
    (ChatGLM2-6B, 1.34)
    (Qwen-14B, 1.15)
    (Qwen-7B, 0.90)
    (Orca-2-7B, 0.57)
    (InternLM-20B, 0.85)
    (Baichuan2-13B-Base, 0.46)
    (Qwen-1.8B, 0.53)
    (Yi-34B, 0.64)
    (Aquila-7B, 0.97)
    (ChatGLM3-6B, 0.47)
    (Yi-6B, 0.57)
    (InternLM-7B, 0.85)
    (Grok-1, 0.01)
    (LLaMA2-7B, 0.04)
    (Mistral-7B-v0.1, 0.01)
    (Gemma-7B, 0.13)
    (InternLM2-7B-Base, -0.12)
    (LLaMA-7B, 0.08)
    (Llama-3-8B,0.16)
    (Baichuan-13B-Base, 0.04)
    (DeepSeekMath-7B, 0.08)
    (Baichuan-7B, 0.03)
    (InternLM2-20B-Base, -0.12)
    (Baichuan2-7B-Base, 0.00)
    (Gemma-2B, 0.06)
    (Phi-1.5, -0.41)
    (Phi-2, -0.07)
    };
    % unseen
    \addplot+[ybar, bar shift=2pt, fill=darkgreen, fill opacity = 0.8, draw=none] coordinates {
    (Aquila2-7B, 2.38)
    (InternLM2-20B, 0.62)
    (InternLM2-7B, 0.72)
    (Aquila2-34B, 0.97)
    (ChatGLM2-6B, 0.83)
    (Qwen-14B, 0.80)
    (Qwen-7B, 0.45)
    (Orca-2-7B, 0.35)
    (InternLM-20B, 0.52)
    (Baichuan2-13B-Base, 0.31)
    (Qwen-1.8B, 0.42)
    (Yi-34B, 0.44)
    (Aquila-7B, 0.86)
    (ChatGLM3-6B, 0.40)
    (Yi-6B, 0.45)
    (InternLM-7B, 0.77)
    (Grok-1, 0.02)
    (LLaMA2-7B, 0.05)
    (Mistral-7B-v0.1, 0.03)
    (Gemma-7B, 0.15)
    (InternLM2-7B-Base, -0.08)
    (LLaMA-7B, 0.11)
    (Baichuan-13B-Base, 0.07)
    (DeepSeekMath-7B, 0.11)
    (Baichuan-7B, 0.07)
    (InternLM2-20B-Base, -0.07)
    (Baichuan2-7B-Base, 0.04)
    (Gemma-2B, 0.11)
    (Phi-1.5, -0.33)
    (Phi-2, 0.00)
    };
  \end{axis}

  \begin{axis}[
      width=14.2cm,
      height=5cm,
      ymin=-15, ymax=150, % Adjust the max value based on your line plot data
      enlarge x limits=0.05,
      symbolic x coords={Aquila2-7B, InternLM2-20B, InternLM2-7B, Aquila2-34B, ChatGLM2-6B, Qwen-14B, Qwen-7B, Orca-2-7B, InternLM-20B, Baichuan2-13B-Base, Qwen-1.8B, Yi-34B, Aquila-7B, ChatGLM3-6B, Yi-6B, InternLM-7B, Grok-1, LLaMA2-7B, Mistral-7B-v0.1,  Gemma-7B, InternLM2-7B-Base, LLaMA-7B, Llama-3-8B, Baichuan-13B-Base, DeepSeekMath-7B, Baichuan-7B, InternLM2-20B-Base, Baichuan2-7B-Base, Gemma-2B, Phi-1.5, Phi-2},
      secondary axis style,
      ylabel={Percentage decrease $\delta$},
      at={(0,0)}, % Position of the first subplot
      ylabel style = {font=\tiny},
      yticklabel shift = -2 pt,
      xticklabel style={font=\tiny},
      yticklabel style={font=\tiny},
      title style={font=\tiny, yshift=-8pt},
      grid style=dashed,
  ]
  % seen
    \addplot[sharp plot, lightgreen, opacity = 0.5, line width=0.7pt, mark=x, mark options={solid, scale=1.0}] coordinates {
    (Aquila2-7B,  141.55)
    (InternLM2-20B, 89.76)
    (InternLM2-7B, 89.66)
    (Aquila2-34B, 87.33)
    (ChatGLM2-6B, 53.17)
    (Qwen-14B, 62.50)
    (Qwen-7B, 42.06)
    (Orca-2-7B, 34.34)
    (InternLM-20B, 34.00)
    (Baichuan2-13B-Base, 28.40)
    (Qwen-1.8B, 33.76)
    (Yi-34B, 26.67)
    (Aquila-7B, 48.02)
    (ChatGLM3-6B, 25.13)
    (Yi-6B, 20.50)
    (InternLM-7B, 29.21)
    (Grok-1,0.22)
    (LLaMA2-7B, 1.11)
    (Mistral-7B-v0.1, 0.31)
    (Gemma-7B, 4.38)
    (InternLM2-7B-Base, -2.60)
    (LLaMA-7B, 2.20)
    (Llama-3-8B,3.94)
    (Baichuan-13B-Base, 1.18)
    (DeepSeekMath-7B, 2.52)
    (Baichuan-7B, 0.80)
    (InternLM2-20B-Base, -2.85)
    (Baichuan2-7B-Base, 0.00)
    (Gemma-2B, 1.73)
    (Phi-1.5, -9.83)
    (Phi-2, -2.03)
    };
    % unseen
    \addplot[sharp plot, darkgreen, opacity = 0.5,  line width=0.7pt, mark=x, mark options={solid, scale=1.0}] coordinates {
    (Aquila2-7B, 66.30)
    (InternLM2-20B, 21.99)
    (InternLM2-7B, 24.91)
    (Aquila2-34B, 47.55)
    (ChatGLM2-6B, 24.41)
    (Qwen-14B, 35.56)
    (Qwen-7B, 16.25)
    (Orca-2-7B, 17.24)
    (InternLM-20B, 17.51)
    (Baichuan2-13B-Base, 16.58)
    (Qwen-1.8B, 23.46)
    (Yi-34B,16.48)
    (Aquila-7B,38.22)
    (ChatGLM3-6B,19.70)
    (Yi-6B,15.20)
    (InternLM-7B,25.08)
    (Grok-1, 0.61)
    (LLaMA2-7B,1.40)
    (Mistral-7B-v0.1,0.94)
    (Gemma-7B,5.08)
    (InternLM2-7B-Base,-1.77)
    (LLaMA-7B,3.05)
    (Llama-3-8B,4.82)
    (Baichuan-13B-Base,2.09)
    (DeepSeekMath-7B,3.49)
    (Baichuan-7B,1.88)
    (InternLM2-20B-Base,-1.69)
    (Baichuan2-7B-Base,1.23)
    (Gemma-2B,3.21)
    (Phi-1.5,-8.09)
    (Phi-2,0.00)
    };
    %  seen- unseen
    \addplot[sharp plot, purple, opacity = 0.8, line width=0.7pt, mark=o, mark options={solid, scale=0.8}] coordinates {
    (Aquila2-7B, 75.25)
    (InternLM2-20B, 67.77)
    (InternLM2-7B, 64.75)
    (Aquila2-34B, 39.78)
    (ChatGLM2-6B, 28.76)
    (Qwen-14B, 26.94)
    (Qwen-7B, 25.81)
    (Orca-2-7B, 17.10)
    (InternLM-20B, 16.49)
    (Baichuan2-13B-Base, 11.82)
    (Qwen-1.8B, 10.30)
    (Yi-34B, 10.19)
    (Aquila-7B, 9.80)
    (ChatGLM3-6B, 5.43)
    (Yi-6B, 5.30)
    (InternLM-7B, 4.13)
    (Grok-1, -0.39)
    (LLaMA2-7B, -0.29)
    (Mistral-7B-v0.1, -0.63)
    (Gemma-7B, -0.70)
    (InternLM2-7B-Base, -0.83)
    (LLaMA-7B, -0.85)
    (Llama-3-8B, -0.88)
    (Baichuan-13B-Base, -0.91)
    (DeepSeekMath-7B, -0.97)
    (Baichuan-7B, -1.08)
    (InternLM2-20B-Base, -1.16)
    (Baichuan2-7B-Base, -1.23)
    (Gemma-2B, -1.48)
    (Phi-1.5, -1.74)
    (Phi-2, -2.03)
    };
  \end{axis}

    \begin{axis}[
        hide axis, 
        xmin=0, xmax=1, ymin=0, ymax=0.1, 
        legend columns=-1, 
        legend style={
            draw=none, 
            at={(1.38, 0.58)}, 
            anchor=north, 
            font=\tiny, 
        }
    ]
    \addlegendimage{ybar,ybar legend,fill=lightgreen,draw=none}
    \addlegendentry{$\Delta_\text{train}$}
    \addlegendimage{ybar,ybar legend,fill=darkgreen,draw=none}
    \addlegendentry{$\Delta_\text{test}$}
    \addlegendimage{sharp plot,lightgreen, mark=x, mark options={solid, scale=1.0}}
    \addlegendentry{$\delta_\text{train}$}
    \addlegendimage{sharp plot,darkgreen, mark=x, mark options={solid, scale=1.0}}
    \addlegendentry{$\delta_\text{test}$}
    \addlegendimage{sharp plot, purple, mark=o, mark options={solid, scale=0.8}}
    \addlegendentry{$\delta_\text{train-test}$}
    \end{axis}
\end{tikzpicture}
\vspace{-15pt}
\caption{LLMs ordered by  $\delta_{\text{train-test}}$  w.r.t ppl on GSM8K.
        The left y-axis indicates the decrement $\Delta$, and the right y-axis shows the percentage decrease $\delta$.  The risk of leakage declines from left to right.  Models positioned on the left are more likely to train on the training set compared to the test set, with this likelihood diminishing as one moves to the right. Findings based on the PPL metric strongly align with those from the 5-gram accuracy.}
 \label{fig:main_results(ppl-G)}
\end{figure}
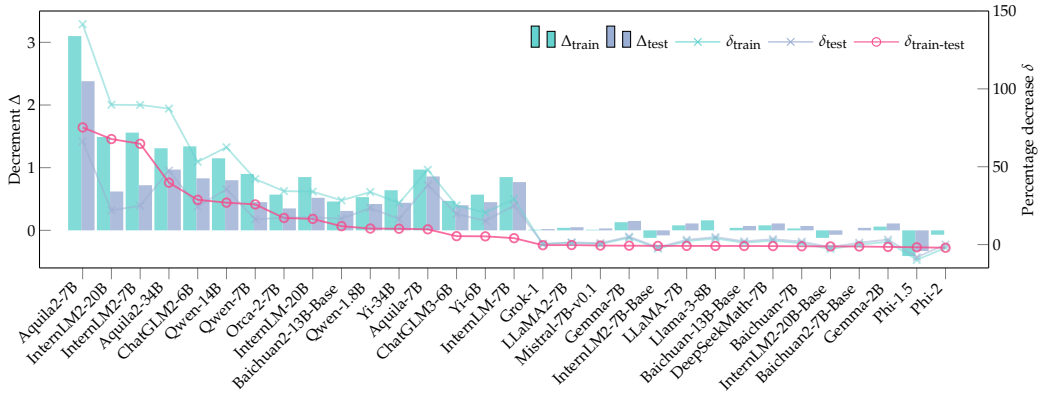

We evaluated 31 LLMs of varying sizes using three atomic metrics—ppl, 5-gram accuracy, and 10-gram accuracy—on the GSM8K and MATH datasets. Comprehensive results are presented in Tables \ref{tab:GSM8K-5gram}-\ref{tab:MATH-ppl}. We visualized the results for 5-gram accuracy and ppl on GSM8K in Figures \ref{fig:main_results(5gram-G)} and \ref{fig:main_results(ppl-G)}, respectively.  The results are ordered by $\delta_\text{train-test}$ scores, which reflect the potential degree of data leakage from the training set relative to the test set. We observe that models such as the LLaMA series and Mistral-7B-v0.1 exhibit nearly zero $\delta_\text{train-test}$ scores, suggesting minimal potential for benchmark leakage. In contrast, the Aquila2 series has been unexpectedly trained on GSM8K data as noted in their documentation,\footnote{\url{https://huggingface.co/BAAI/AquilaChat2-34B/blob/main/README.md}} and InternLM-2 (excluding the Base version) has undergone continual pre-training on STEM data, including suspected exposure to the GSM8K training set according to its technical report~\citep{cai2024internlm2}. These factors align with their significantly higher  $\delta_\text{train-test}$  scores from our detection pipeline, indicating potential benchmark data utilization in both the Aquila2 series and InternLM-2 series (excluding the Base version). Meanwhile, the models located near these in the Figure~\ref{fig:main_results(5gram-G)} and Figure~\ref{fig:main_results(ppl-G)} might harbor a risk of data leakage. 

Additionally, it should be noted that when a small absolute $\Delta$ value is paired with a large percentage, it suggests that even slight variations lead to substantial percentage fluctuations, thus rendering the $\delta_{\text{train-test}}$ less reliable, as exemplified by Phi-2 (cf. Figure~\ref{fig:main_results(ppl-G)}). It is important to note that even if the $\delta_\text{train-test}$ scores of some models are not significant, this does not rule out the possibility of data leakage. It is possible that both the training and test sets have been contaminated, which could lead to undetectably low $\delta_\text{train-test}$ scores. In such cases, we can supplement our observations with the model's  $\Delta$ and $\delta$ scores on the training and test sets (also n-gram prediction observations from \S~\ref{sec:instance-level-dection}), determining if they are both relatively high on the train and test datasets.

When combining Figure~\ref{fig:main_results(5gram-G)} with Figure~\ref{fig:main_results(ppl-G)}, we can find that the top-ranked models in both figures are essentially the same, including Aquila2, InternLM-2, and Qwen series, despite the order slightly varying. Similar observations can be seen in the MATH dataset, as evidenced by results presented in Tables~\ref{tab: MATH-5gram}, \ref{tab: MATH-10gram}, and \ref{tab:MATH-ppl}. The discrepancies observed in rankings under the two metrics-n-gram accuracy and ppl can be attributed to their fundamentally different approaches to language modeling. Perplexity provides a continuous measure, capturing probabilistic nuances across a broad range of text samples. In contrast, n-gram accuracy requires exact matches and is thus more sensitive to the organizational format of data. This characteristic makes it less adept at recognizing subtleties in cases such as paraphrasing or reformatting within benchmark datasets. For example, our closer observation on predictions revealed that when predicting 5-grams near the answer, the model ChatGLM-2 often predicts    ``\texttt{Answer: }\verb|\|\verb|\|\texttt{boxed}'', whereas the golden 5-gram is ``\verb|\|\texttt{n\#\#\# 12}''. This discrepancy suggests that the model may have been trained on a rephrased or reformatted version of the GSM8K benchmark dataset. This insight supports the notion that while n-gram accuracy fails to capture such nuanced training effects, perplexity can provide a broader understanding of a model's adaptation to modified data inputs.

\subsection{N-gram Accuracy Helps Instance-level Leakage Detection}
\label{sec:instance-level-dection}

\begin{figure}[h]
    \centering
    \includegraphics[width=\textwidth]{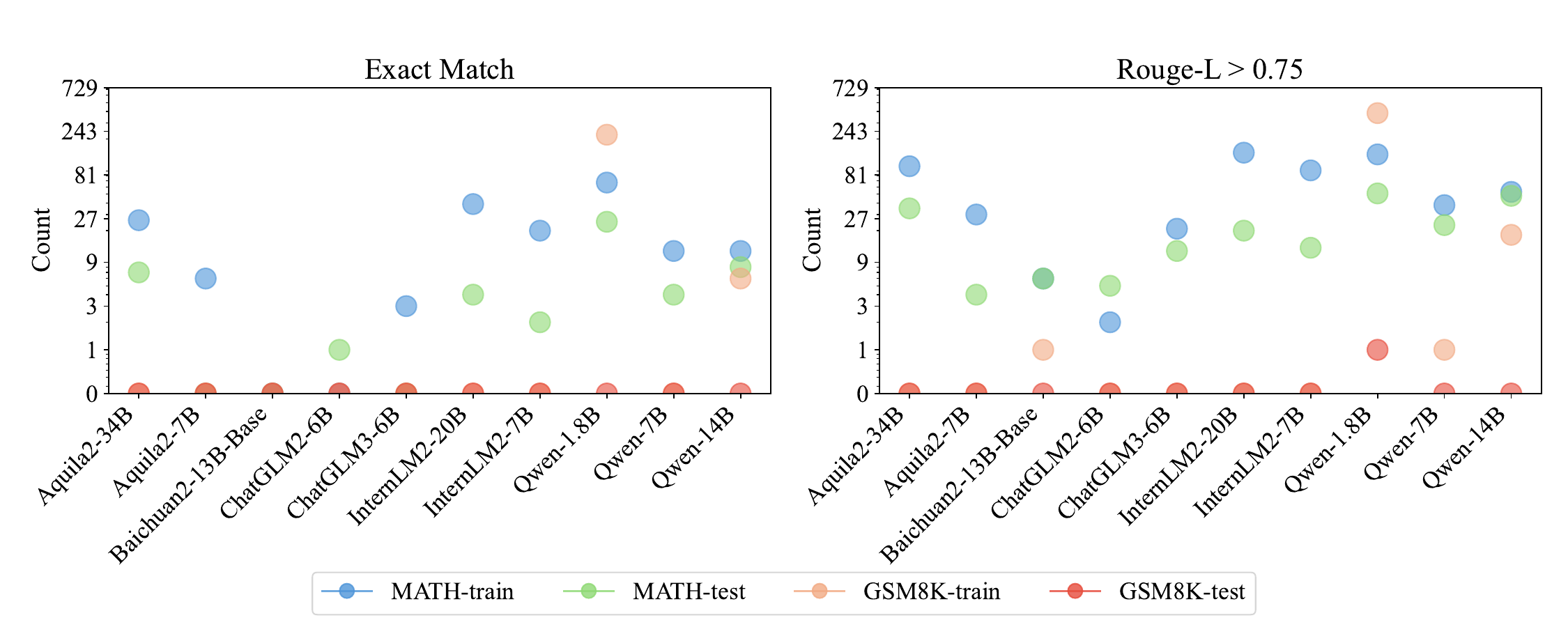}
        \caption{Statistics of suspicious leaked sample, where all 5-grams within a sample are predicted correctly, either strictly (measured by Exact Match) or loosely (measured by ROUGE-L). The y-axis employs an exponential scale based on powers of 3.}
    \label{fig:instance-statistics}
\end{figure}

As previously mentioned, and illustrated in Figure~\ref{fig:overview} (b) high accuracy for each n-gram of an example's prediction suggests a high probability that the sample was encountered during the training process. To investigate instance-level leakage, we looked closer at n-gram predictions across different models. Additionally, considering that benchmark data may undergo reformatting, paraphrasing, or other modifications when integrated into model training, we leverage lenient metrics, such as ROUGE-L and edit distance similarity (cf. \S~\ref{appendix-sec:edit_distance}), for comparing n-grams. Under this context, an instance is deemed correctly predicted if it achieves an Exact Match (meaning all predictions align perfectly), or if the edit distance similarity of all predictions exceeds 0.9 (indicating substantial similarity), and further, if the ROUGE-L score of all predictions surpasses 0.75. 

We compiled statistics across various stringent-to-lenient scoring thresholds to quantify instances where all 5-grams were accurately predicted by several models, with part of the results shown in Figure~\ref{fig:instance-statistics} and detailed results presented in the Table~\ref{tab:Exact-Match-Instance-Statistics}-\ref{tab:Rouge-L-Instance-Statistics}. Surprisingly, Qwen-1\_8B can precisely replicate many n-gram predictions from the MATH and GSM8K datasets. Specifically, it accurately predicted all 5-grams in 223 examples from the GSM8K training set and 67 from the MATH training set, with an additional 25 correct predictions even in the MATH test set. These observations complement the results discussed in \S~\ref{sec:main-results}, where Qwen-1\_8B may not rank highest in $\delta_{\text{train-test}}$ scores but exhibits high $\Delta_\text{train}$ and $\delta_\text{train}$ scores (also high $\Delta_\text{test}$ and $\delta_\text{test}$ scores), as shown in Figure~\ref{fig:main_results(5gram-G)}. This aligns perfectly with our observation that it can accurately and completely replicate many n-grams from the training set. We would like to emphasize that the n-gram accuracy metric can mitigate issues in our detection pipeline, particularly when the training and test datasets are simultaneously leaked and remain undetected. However, this also has its limitations; it can only detect examples that are integrated into the model training in their original format and wording, unless we know the organizational format of the training data used by the model in advance.

\subsection{Case Study}

\begin{figure}[h]
    \centering
    \includegraphics[width=\textwidth]{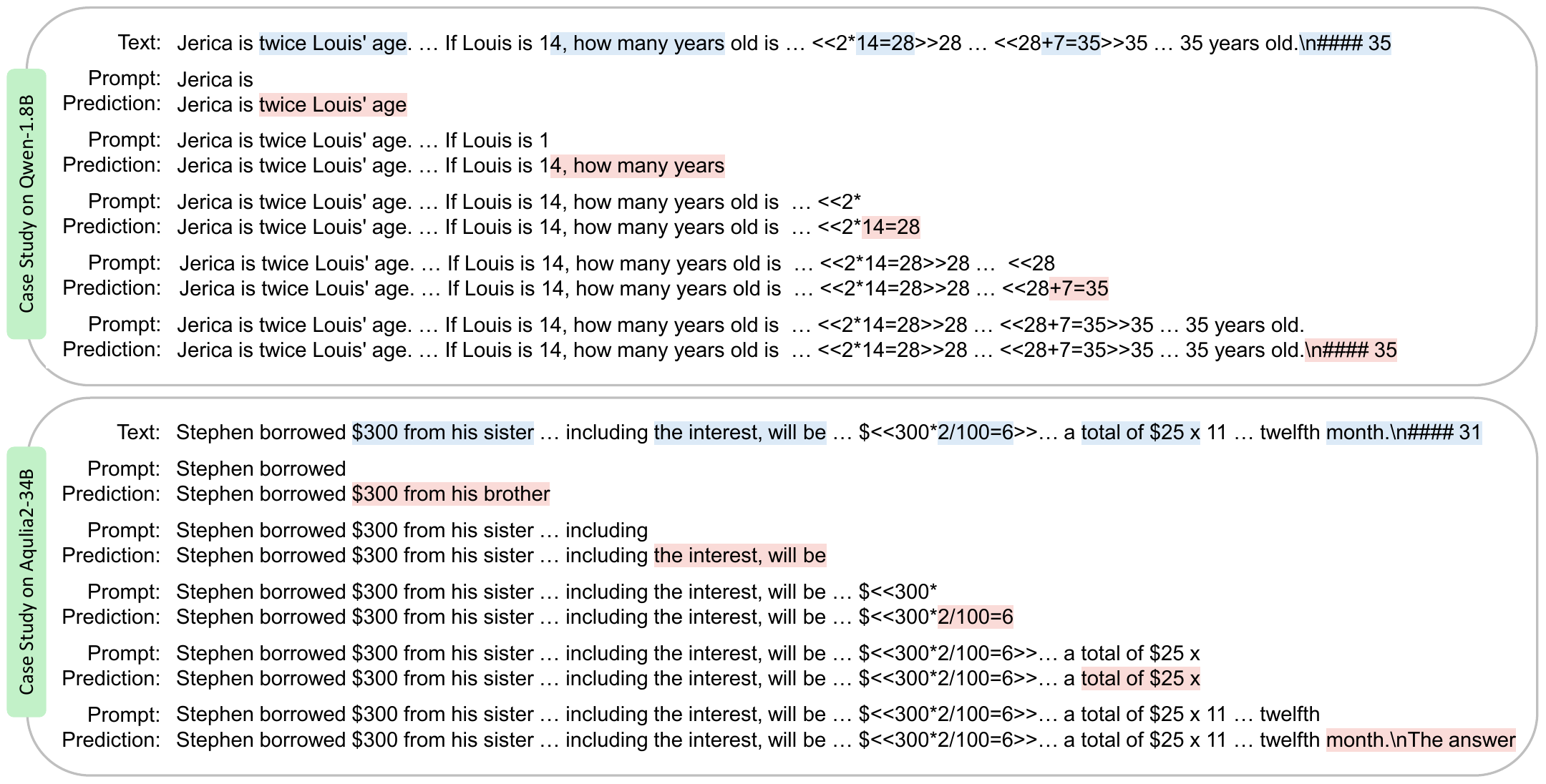}
        \caption{      
Two cases: one from the GSM8K training set predicted by the Qwen-1.8B model (above), and one from the GSM8K test set by the Aquila2-34B model (below). Both examples are presented with the original question and answer concatenated, separated by a space.}
    \label{fig:instance_case}
\end{figure}

To gain a deeper understanding of model behaviors, we took a closer look at 
the n-gram predictions and highlight some cases in Figure~\ref{fig:instance_case}. In the first case, the Qwen-1.8B model achieves perfect n-gram predictions on a sample from the GSM8K training set, completing all 5-grams accurately. This strongly suggests potential data leakage within the training set of GSM8K. Additionally, we also conducted a case study on the Aquila2-34B model, known to accidentally be exposed to the entire GSM8K test set. It consistently predicts n-grams as  ``The answer is'' for all instances where the ground truth was represented by a placeholder "\#\#\#\#". This observation exactly explains why it is challenging to detect  leakage using our n-gram accuracy metric.  Consequently, the presence of a high $\delta_{\text{train-test}}$ score, coupled with poor n-gram accuracy, hints that the model might have undergone data augmentation or reformatting during its training process. To enhance readers' comprehension of model behaviors, we have released an interactive demo for case studies.

\section{Recommendation for Model Documentation and Benchmarks Setup}
\label{sec: recommendation}

To ensure fairness in the evaluation of large language models moving forward, we propose the following suggestions: 

\begin{itemize}
    \item Model Documentation: For any LLMs to be released, comprehensive documentation should be provided. This documentation at least specifies \textbf{whether the model has been trained on the training or test sets of commonly used benchmarks to prevent potentially unfair comparisons}. To this end, we introduce \emph{Benchmark Transparency Card} (cf. Table~\ref{tab:benchmark-transparency-card}), which serves as the supplement of the Data Card~\citep{10.1145/3531146.3533231-data-cards,10.1145/3458723-datasheets-for-datasets} and Model Card~\citep{10.1145/3287560.3287596-model-cards}, aiming to document the utilization of benchmarks (such as whether any benchmark sets are used for training and whether any data augmentation techniques are applied) and benchmark evaluation details (cf. \S~\ref{sec:benchmark-transparency-card} for more details). We hope that this card will be widely adopted upon the release of models to foster the healthy development of large language models.
    \item  Benchmark Construction: We recommend constructing benchmarks from the latest corpus to minimize the risk of overlap with pre-training corpora. Additionally, evaluation datasets should be regularly updated using dynamic benchmarks to guard against overfitting to static test datasets~\citep{zhu2023dyval,zhu2024dyval2,jain2024livecodebench}. 
    \item  Benchmark Public Access: To mitigate the risk of \texttt{Input-Output Leakage}, we advise against directly uploading original benchmarks online, particularly when they contain paired questions and answers. As suggested by \cite{Jacovi2023Stop-upload-testset}, encrypting the test set prior to uploading can enhance security. Alternatively, maintaining a private test set through a leaderboard format is also a viable option. 
    \item Evaluation: We recommend caution in drawing overly optimistic conclusions about a model's capabilities based on its strong performance in specific benchmarks. It may be beneficial to evaluate the model further using a variety of contemporary challenges, such as new exam questions~\citep{testing_language_models_on_a_held_out_high_school_national_finals_exam}, to provide a more balanced assessment of its abilities. When benchmarking proprietary models, it is important to proceed with caution, especially when submitting benchmark data through APIs. There is a risk that this data could be utilized by the model's provider for further training purposes.
\end{itemize}

\section{Related Work}

\noindent\textbf{Benchmarks in Natural Language Processing} \quad As LLMs grow in capability, rigorous evaluations are crucial for tracking progress. Historically, benchmark datasets like the Penn Treebank~\citep{marcus1993building}, SST~\citep{socher-etal-2013-recursive}, SQuAD~\citep{rajpurkar-etal-2016-squad}, GLUE~\citep{wang-etal-2018-glue}, SuperGLUE~\citep{DBLP:conf/nips/WangPNSMHLB19-superglue}, and GEM~\citep{gehrmann-etal-2021-gem} have been pivotal for NLP advancements. With the trend of scaling up models, there's been an enhancement in model capabilities, leading to the development of specialized evaluation suites by \citet{DBLP:conf/nips/BrownMRSKDNSSAA20-gpt3}, world knowledge assessment MMLU~\citep{DBLP:conf/iclr/HendrycksBBZMSS21-mmlu}, CMMLU~\citep{DBLP:journals/corr/abs-2306-09212-cmmlu}, and C-Eval~\citep{huang2023ceval}, and diverse benchmarks like BIG-bench~\citep{DBLP:journals/corr/abs-2206-04615-bigbench}. Other efforts include the EleutherAI LM Harness for few-shot evaluation~\citep{lm-eval-harness}, \citet{liang2023holistic}'s holistic evaluation framework, AGIEval for human-centric standardized exams~\citep{DBLP:journals/corr/abs-2304-06364-AGIEval}, and conversational model benchmarks like MT-Bench and ChatbotArena~\citep{DBLP:journals/corr/abs-2306-05685-mtbench}. In this work, we primarily focus on challenging mathematical reasoning tasks such as GSM8K~\citep{DBLP:journals/corr/abs-2110-14168-gsm8k} and MATH~\citep{DBLP:conf/nips/HendrycksBKABTS21-math} as our test bed for leak detection.

\noindent\textbf{Data Leakage Detection}  \quad As pre-training datasets grow, the inadvertent inclusion of benchmark data into the training corpus becomes more likely. Early studies, such as those on GPT-2\citep{radford2019gpt2}, GPT-3~\citep{DBLP:conf/nips/BrownMRSKDNSSAA20-gpt3}, FLAN~\citep{DBLP:conf/iclr/WeiBZGYLDDL22-FLAN} and LLaMA-2~\citep{DBLP:journals/corr/abs-2307-09288-llama-2}, utilized post-hoc n-gram overlap analyses between benchmarks and pre-training corpora to gauge data leakage. Access to pre-training data enables leakage detection through overlap analysis, as highlighted in previous research~\citep{dodge-etal-2021-documenting,DBLP:journals/corr/abs-2311-09783-investigating-data-contamination-allenai}; even without direct access, details like data sources and time frames can reveal leakages~\citep{DBLP:journals/corr/abs-2309-10677-estimating-contamination-via-ppl,Li2023-open-source-data-contamination-report}. However, as model training becomes less transparent, identifying leaks poses greater challenges~\citep{shi2024detecting}. Our work addresses scenarios lacking access to pre-training data or training details, striving for transparency in large model development. Researchers have developed various detection methods, including benchmark perturbations~\citep{DBLP:journals/corr/abs-2310-17623-prove-test-set-contamination,DBLP:journals/corr/abs-2311-09783-investigating-data-contamination-allenai} and leveraging synthetic data, akin to efforts by \citet{DBLP:journals/corr/abs-2310-19341-skywork} in estimating leakage using a dataset similar to GSM8K, despite potential data drift risks. Paraphrasing benchmarks, as seen in the work of \citet{DBLP:journals/corr/abs-2311-09154-clean-eval,DBLP:journals/corr/abs-2311-06233-data-contamination-quiz}, shares similarities with our approach. Additionally, studies simulating data leakage by blending benchmarks into training data explore its performance impact~\citep{DBLP:journals/corr/abs-2311-01964-benchmark-cheater, DBLP:journals/corr/abs-2401-06059-investigating-data-contamination}. These investigations underline the importance of understanding models' data memorization, an area explored by prior research~\citep{DBLP:conf/uss/CarliniTWJHLRBS21-extracting-training-data,DBLP:conf/acl/Magar022-memorization-to-exploitation}. Our method, independent of instruction-following capabilities, utilizes basic language modeling metrics and data synthesis for detecting data leakage, even achieving instance-level granularity.

\section{Conclusion and Limitations}

In this study, we first summarize various typical training behaviors and challenges associated with detecting benchmark leakage. We then introduce a simple, computationally efficient, and scalable pipeline based on two fundamental metrics for leakage detection, validated through meta-experiments. The n-gram accuracy metric we introduce can be utilized for instance-level detection, capable of accurately identifying many instances that have been trained.  Our comprehensive analysis of 31 open-source LLMs unveils that many may have exploited benchmark data to enhance their performance on mathematical reasoning tasks, thereby gaining an unfair advantage. Additionally, we offer several recommendations from diverse perspectives. Notably, we advocate for the adoption of the ``\emph{Benchmark Transparency Card}'' to promote transparent documentation of benchmark usage, fostering more ethical development of LLMs. However, this work has limitations. For example, our method might not detect cases where models are trained on benchmarks that have been augmented or reformatted, and where there is concurrent leakage in training and testing datasets. Despite these challenges, our n-gram accuracy metric mitigates some of these issues.

\section{Acknowledgements}
We particularly thank Ethan Chern, Yan Ma, Haoyang Zou, Xuefeng Li, Yiyuan Li, and other lab mates for their helpful and detailed suggestions.

\bibliography{colm2024_conference}
\bibliographystyle{colm2024_conference}

\clearpage

\appendix
\section{Appendix}

\subsection{Reference Benchmark Synthesis}

We provide the full prompts for data synthesis in Table~\ref{tab:paraphrase_prompt_gsm8k} and Table~\ref{tab:paraphrase_prompt_math} and synthesized examples in Table~\ref{tab:paraphrase_example_GSM8K} and Table~\ref{tab:paraphrase_example_math}.

\subsection{The Details of Meta Experiments}
\label{appendix-sec:meta-experiments}

For details for meta experiments, specifically, we train Mistral-7B-v0.1 for 20 epochs on GSM8K and 8 epochs on MATH, using the AdamW optimizer with a sequence length of 8,192 tokens. The batch size is 64 for those two datasets. The AdamW optimizer's hyperparameters are set as follows: $\beta_1=0.9$, $\beta_2=0.95$, $\epsilon=10^{-5}$, and weight decay of $0.1$. We employ a cosine learning rate schedule with a maximum learning rate of $2 \times 10^{-5}$ for the GSM8K dataset with full loss and $1.6 \times 10^{-5}$ with answer loss, and $1 \times 10^{-5}$ for the MATH dataset with full loss and $8.7 \times 10^{-6}$ with answer loss, which decays to 10\% of the maximum value. Following \citet{DBLP:journals/jmlr/GranziolZ022}, \citet{wang2023openchat}, and \citet{fan2024reformatted}, the learning rate is scaled proportionally to the square root of the batch size. All models are trained on 8 NVIDIA A100 80G GPUs.

\subsection{The Description of Edit Distance Similarity}
\label{appendix-sec:edit_distance}
The edit distance similarity is calculated based on edit distance by the following equation:
\begin{equation}
    DS(A, B) = \frac{D(A, B)}{\max(|A|, |B|)},
\end{equation}
where the $DS(A, B)$ is the edit distance similarity between string $A$ and $B$, the $D(A, B)$ is the edit distance between string $A$ and $B$, and $|A|$ is the length of string $A$.

The edit distance between two strings is a measure of their dissimilarity, quantifying how many operations (i.e., insertions, deletions, or substitutions) are required to transform one string into the other. The algorithm to calculate this distance is based on dynamic programming, known as the Levenshtein distance.

Given two strings, \(A\) of length \(m\) and \(B\) of length \(n\), we define \(D(i, j)\) as the edit distance between the first \(i\) characters of \(A\) and the first \(j\) characters of \(B\). The edit distance between \(A\) and \(B\) can be computed using the following recurrence relation:

\begin{equation}
D(i, j) = 
\begin{cases} 
i & \text{if } j = 0, \\
j & \text{if } i = 0, \\
D(i-1, j-1) & \text{if } A[i] = B[j], \\
1 + \min\{D(i-1, j), D(i, j-1), D(i-1, j-1)\} & \text{otherwise.}
\end{cases},
\end{equation}
where \(D(i-1, j)\) represents the distance after an insertion, \(D(i, j-1)\) corresponds to the distance after a deletion, and \(D(i-1, j-1)\) accounts for a substitution if \(A[i] \neq B[j]\), or matches \(A[i]\) with \(B[j]\) if they are equal.

To initialize, \(D(0, 0) = 0\) since no operations are needed to transform an empty string into another empty string. The values of \(D(i, 0)\) for \(i=1 \ldots m\) and \(D(0, j)\) for \(j=1 \ldots n\) are initialized to \(i\) and \(j\) respectively, representing the distance of transforming a string of length \(i\) or \(j\) into an empty string through deletions. This dynamic programming approach efficiently computes the edit distance by building up the solution using previously computed sub-problems. The final edit distance is given by \(D(m, n)\), corresponding to the distance between the full lengths of \(A\) and \(B\). During implementation, the edit distance was computed with the ``\texttt{editdistance}'' pip package.\footnote{\url{https://pypi.org/project/editdistance/}}

\begin{table}[ht]
    \scriptsize
    \centering
\begin{tabular}{@{}p{1\columnwidth}@{}}
\toprule
\textbf{System Prompt}\\
Please act as a mathematics problem rewriter to paraphrase the problem and the answer presented below.

\\\\
Please follow the instructions below:

1. Please paraphrase the problem by rewording it with new expressions and sentence structures.

2. Please do not change the essence of the problem and the answer.

3. Please make sure not to deviate too much from the original content, and try to maintain the same style as much as possible.

4. Please imitate the original answer and output the final answer in the last line using \#\#\#\#, containing only numbers. 

\\\\
Please write "The rewritten question: $<$question$>$" to output your rewritten question without any additional information, and write "The rewritten answer: $<$answer$>$" to output your rewritten answer without any additional information.

\\\\
There is an example for your reference:

\\\\
Question: Weng earns \$12 an hour for babysitting. Yesterday, she just did 50 minutes of babysitting. How much did she earn?

\\\\

Answer: Weng earns 12/60 = \$$<<$12/60=0.2$>>$0.2 per minute.$\backslash$nWorking 50 minutes, she earned 0.2 $\times$ 50 = \$$<<$0.2*50=10$>>$10.$\backslash$n\#\#\#\# 10

\\\\
The rewritten question: Weng is paid \$12 per hour for her babysitting services. If she spent 50 minutes babysitting yesterday, what was her total earnings?

\\\\
The rewritten answer: Weng's rate is 12/60 = \$$<<$12/60=0.2$>>$0.2 for every minute. Thus, for 50 minutes of work, she earned 0.2 $\times$ 50 = \$$<<$0.2*50=10$>>$10.$\backslash$n\#\#\#\# 10
\\
\midrule
\textbf{User Prompt}\\
Below is a question and the answer:
\\\\

[Question start]

\{question\}

[Question end]

\\\\

[Answer start]

\{answer\}

[Answer end]
 \\
\bottomrule
\end{tabular}
    \caption{Paraphrasing prompts for GSM8K.}
    \label{tab:paraphrase_prompt_gsm8k}
\end{table}

\begin{table}[t]
    \scriptsize
    \centering
\begin{tabular}{@{}p{1\columnwidth}@{}}
\toprule
\textbf{System Prompt}\\
Please act as a mathematics problem rewriter to paraphrase the problem and the answer presented below.

\\\\
Please follow the instructions below:

1. Please paraphrase the problem by rewording it with new expressions and sentence structures.

2. Please do not change the essence of the problem and the answer.

3. Please make sure not to deviate too much from the original content, and try to maintain the same style as much as possible.

4. Please copy [asy], [/asy], and the code contained within them in its entirety.

\\\\
Please write "The rewritten question: $<$question$>$" to output your rewritten question without any additional information, and write "The rewritten answer: $<$answer$>$" to output your rewritten answer without any additional information.

\\\\
There is an example for your reference:

\\\\
Question: If the system of equations  $\backslash$begin\{align*\}$\backslash$n3x+y\&=a,$\backslash$$\backslash$$\backslash$n2x+5y\&=2a,$\backslash$n$\backslash$end\{align*\} has a solution \$(x,y)\$ when \$x=2\$, compute \$a\$.

\\\\
Answer: Substituting in \$x=2\$, we obtain the equationsn$\backslash$n$\backslash$begin{align*}$\backslash$ny+6\&=a,$\backslash$$\backslash$$\backslash$n5y+4\&=2a.$\backslash$n$\backslash$end\{align*\}$\backslash$n$\backslash$nMultiplying the first equation by \$5\$ and subtracting it from the second equation, we find$\backslash$n$\backslash$n\$\$-26=-3a$\backslash$Rightarrow a=$\backslash$boxed\{$\backslash$frac\{26\}\{3\}\}.\$\$

\\\\
The rewritten question: Examine if the pair of equations given below has a solution \$(x,y)\$ where \$x=2\$, then determine the value of \$a\$. $\backslash$n$\backslash$n$\backslash$begin\{align*\}$\backslash$n3x+y\&=a,$\backslash$$\backslash$$\backslash$n2x+5y\&=2a,$\backslash$n$\backslash$end\{align*\}

\\\\
The rewritten answer: By inserting \$x=2\$ into the equations, we get: $\backslash$n$\backslash$n$\backslash$begin\{align*\}$\backslash$ny+6\&=a,$\backslash$$\backslash$$\backslash$n5y+4\&=2a.$\backslash$n$\backslash$end\{align*\} $\backslash$n$\backslash$nThen, by multiplying the initial equation by \$5\$ and deducting from the second, we ascertain:$\backslash$n$\backslash$n\$\$-26=-3a$\backslash$Rightarrow a=$\backslash$boxed\{$\backslash$frac\{26\}\{3\}\}.\$\$
\\
\midrule
\textbf{User Prompt}\\
Below is a question and the answer:
\\\\

[Question start]

\{question\}

[Question end]

\\\\

[Answer start]

\{answer\}

[Answer end]
 \\
\bottomrule
\end{tabular}
    \caption{Paraphrasing prompts for MATH.}
    \label{tab:paraphrase_prompt_math}
\end{table}

\begin{table}[t]
    \scriptsize
    \centering
\begin{tabular}{@{}p{1\columnwidth}@{}}
\toprule
\textbf{Original Question}\\
Hana sold 4/7 of her stamp collection for \$28. How much would she have earned from selling the entire collection?
\\
\midrule
\textbf{Original Answer}\\
Hana sold 4/7 of her collection for \$28, so 1/7 of her collection represents: 28/4 = \$$<<28/4=7>>7$.$\backslash$nAnd as a result, the entire collection represents: 7 * 7 = \$$<<7*7=49>>49$.$\backslash$n\#\#\#\# 49
\\
\midrule
\textbf{Paraphrased Question}\\
Hana received \$28 by selling 4/7 of her stamp collection. How much money would she have made if she sold the entire collection?
\\
\midrule
\textbf{Paraphrased Answer}\\
If Hana sold 4/7 of her collection for \$28, then 1/7 of her collection is worth: 28/4 = \$$<<28/4=7>>7$. Consequently, the total value of her entire collection is: 7 * 7 = \$$<<7*7=49>>49$.$\backslash$n\#\#\#\# 49
 \\
\bottomrule
\end{tabular}
    \caption{An example of paraphrased data in GSM8K.}
    \label{tab:paraphrase_example_GSM8K}
\end{table}

\begin{table}[t]
    \scriptsize
    \centering
\begin{tabular}{@{}p{1\columnwidth}@{}}
\toprule
\textbf{Original Question}\\
If the system of equations  $\backslash$begin\{align*\}$\backslash$n3x+y\&=a,$\backslash$$\backslash$$\backslash$n2x+5y\&=2a,$\backslash$n$\backslash$end\{align*\} has a solution \$(x,y)\$ when \$x=2\$, compute \$a\$.
\\
\midrule
\textbf{Original Answer}\\
Substituting in \$x=2\$, we obtain the equations$\backslash$n$\backslash$n$\backslash$begin\{align*\}$\backslash$ny+6\&=a,$\backslash$$\backslash$$\backslash$n5y+4\&=2a.$\backslash$n$\backslash$end\{align*\}$\backslash$n$\backslash$nMultiplying the first equation by \$5\$ and subtracting it from the second equation, we find$\backslash$n$\backslash$n\$\$-26=-3a$\backslash$Rightarrow a=$\backslash$boxed\{$\backslash$frac\{26\}\{3\}\}.\$\$
\\
\midrule
\textbf{Paraphrased Question}\\
Determine the value of \$a\$ if the set of equations provided below has a solution \$(x,y)\$ for \$x=2\$. $\backslash$n$\backslash$n$\backslash$begin\{align*\}$\backslash$n3x+y\&=a,$\backslash$$\backslash$$\backslash$n2x+5y\&=2a,$\backslash$n$\backslash$end\{align*\}
\\
\midrule
\textbf{Paraphrased Answer}\\
Upon substituting \$x=2\$ into the equations, we derive:$\backslash$n$\backslash$n$\backslash$begin\{align*\}$\backslash$ny+6\&=a,$\backslash$$\backslash$$\backslash$n5y+4\&=2a.$\backslash$n$\backslash$end\{align*\}$\backslash$n$\backslash$nBy multiplying the first equation by \$5\$ and subtracting it from the second equation, we establish:$\backslash$n$\backslash$n\$\$-26=-3a$\backslash$Rightarrow a=$\backslash$boxed\{$\backslash$frac\{26\}\{3\}\}.\$\$
 \\
\bottomrule
\end{tabular}
    \caption{An example of paraphrased data in MATH.}
    \label{tab:paraphrase_example_math}
\end{table}

%Please add the following packages if necessary:
%\usepackage{booktabs, multirow} % for borders and merged ranges
%\usepackage{soul}% for underlines
%\usepackage[table]{xcolor} % for cell colors
%\usepackage{changepage,threeparttable} % for wide tables
%If the table is too wide, replace \begin{table}[!htp]...\end{table} with
%\begin{adjustwidth}{-2.5 cm}{-2.5 cm}\centering\begin{threeparttable}[!htb]...\end{threeparttable}\end{adjustwidth}
\begin{table}[!htp]
\centering
\scriptsize
\scalebox{0.65}{
\begin{tabular}{lrrrrrrr|rrrrrrr|r}
\toprule
Model & $\mathcal{D}_{\text{seen-ref}_1}$ & $\mathcal{D}_{\text{seen-ref}_2}$ & $\mathcal{D}_{\text{seen-ref}_3}$ & $\overline{\mathcal{D}_{\text{seen-ref}_i}}$ & $\mathcal{D}_{\text{seen}}$ & $\Delta_{\text{seen}}$ & $\delta_{\text{seen}}$ & 
$\mathcal{D}_{\text{unseen-ref}_1}$ & $\mathcal{D}_{\text{unseen-ref}_2}$ & $\mathcal{D}_{\text{unseen-ref}_3}$ & $\overline{\mathcal{D}_{\text{unseen-ref}_i}}$ & $\mathcal{D}_{\text{unseen}}$ &$\Delta_{\text{unseen}}$ & $\delta_{\text{unseen}}$ &\textbf{$\delta_{\text{seen-unseen}}$} \\\midrule
Mistral-7B &15.46 &15.86 &15.98 &15.77 &17.06 &1.29 &7.56 &16.08 &16.12 &16.32 &16.17 &16.62 &0.45 &2.71 &4.85 \\
+ SFT &39.76 &39.86 &39.18 &39.6 &48.22 &8.62 &17.88 &38.32 &38.18 &38.22 &38.24 &42.08 &3.84 &9.13 &8.75 \\
+ Pretrain &40.34 &40.5 &40.04 &40.29 &50.62 &10.33 &20.41 &37.84 &38.8 &38.3 &38.31 &43.52 &5.21 &11.97 &8.44 \\
\bottomrule
\end{tabular}}
\caption{Meta experiment results w.r.t 5-gram accuracy on GSM8K. $\mathcal{D}_\text{seen}$ denotes the sampled set from GSM8K used for training, while $\mathcal{D}_\text{unseen}$ represents the sampled unseen set for the model used for evaluation. $\mathcal{D}_\text{xxx-ref}$ stands for the synthesized referenced datasets.}
\label{tab: Meta-G-5gram}
\end{table}
%Please add the following packages if necessary:
%\usepackage{booktabs, multirow} % for borders and merged ranges
%\usepackage{soul}% for underlines
%\usepackage[table]{xcolor} % for cell colors
%\usepackage{changepage,threeparttable} % for wide tables
%If the table is too wide, replace \begin{table}[!htp]...\end{table} with
%\begin{adjustwidth}{-2.5 cm}{-2.5 cm}\centering\begin{threeparttable}[!htb]...\end{threeparttable}\end{adjustwidth}
\begin{table}[!htp]\centering
\scriptsize
\scalebox{0.65}{
\begin{tabular}{lrrrrrrr|rrrrrrr|r}
\toprule
Model & $\mathcal{D}_{\text{seen-ref}_1}$ & $\mathcal{D}_{\text{seen-ref}_2}$ & $\mathcal{D}_{\text{seen-ref}_3}$ & $\overline{\mathcal{D}_{\text{seen-ref}_i}}$ & $\mathcal{D}_{\text{seen}}$ & $\Delta_{\text{seen}}$ & $\delta_{\text{seen}}$ & 
$\mathcal{D}_{\text{unseen-ref}_1}$ & $\mathcal{D}_{\text{unseen-ref}_2}$ & $\mathcal{D}_{\text{unseen-ref}_3}$ & $\overline{\mathcal{D}_{\text{unseen-ref}_i}}$ & $\mathcal{D}_{\text{unseen}}$ &$\Delta_{\text{unseen}}$ & $\delta_{\text{unseen}}$ &\textbf{$\delta_{\text{seen-unseen}}$} \\\midrule
Mistral-7B &29.14 &31.36 &31.56 &30.69 &33.24 &2.55 &7.67 &28.5 &29.58 &29.5 &29.19 &31.64 &2.45 &7.74 &-0.07 \\
+ SFT &39.3 &40 &41.2 &40.17 &49.04 &8.87 &18.09 &30.5 &31.58 &31.78 &31.29 &34.02 &2.73 &8.02 &10.07 \\
+ Pretrain &39.84 &41.82 &42.68 &41.45 &51.28 &9.83 &19.17 &30.96 &32.04 &32.28 &31.76 &35.66 &3.9 &10.94 &8.23 \\
\bottomrule
\end{tabular}}
\caption{Meta experiment results w.r.t 5-gram accuracy on MATH. $\mathcal{D}_\text{seen}$ denotes the sampled set from MATH used for training, while $\mathcal{D}_\text{unseen}$ represents the sampled unseen set for the model used for evaluation. $\mathcal{D}_\text{xxx-ref}$ stands for the synthesized referenced datasets.}
\label{tab: Meta-M-5gram}
\end{table}
%Please add the following packages if necessary:
%\usepackage{booktabs, multirow} % for borders and merged ranges
%\usepackage{soul}% for underlines
%\usepackage[table]{xcolor} % for cell colors
%\usepackage{changepage,threeparttable} % for wide tables
%If the table is too wide, replace \begin{table}[!htp]...\end{table} with
%\begin{adjustwidth}{-2.5 cm}{-2.5 cm}\centering\begin{threeparttable}[!htb]...\end{threeparttable}\end{adjustwidth}
\begin{table}[!htp]\centering
\scriptsize
\scalebox{0.65}{
\begin{tabular}{lrrrrrrr|rrrrrrr|r}
\toprule
Model & $\mathcal{D}_{\text{seen-ref}_1}$ & $\mathcal{D}_{\text{seen-ref}_2}$ & $\mathcal{D}_{\text{seen-ref}_3}$ & $\overline{\mathcal{D}_{\text{seen-ref}_i}}$ & $\mathcal{D}_{\text{seen}}$ & $\Delta_{\text{seen}}$ & $\delta_{\text{seen}}$ & 
$\mathcal{D}_{\text{unseen-ref}_1}$ & $\mathcal{D}_{\text{unseen-ref}_2}$ & $\mathcal{D}_{\text{unseen-ref}_3}$ & $\overline{\mathcal{D}_{\text{unseen-ref}_i}}$ & $\mathcal{D}_{\text{unseen}}$ &$\Delta_{\text{unseen}}$ & $\delta_{\text{unseen}}$ &\textbf{$\delta_{\text{seen-unseen}}$} \\\midrule
Mistral-7B &3.25 &3.26 &3.22 &3.24 &3.25 &-0.01 &-0.31 &3.22 &3.21 &3.22 &3.22 &3.2 &0.02 &0.63 &-0.94 \\
+ SFT &1.85 &1.86 &1.85 &1.85 &1.31 &0.54 &41.22 &1.95 &1.94 &1.95 &1.95 &1.6 &0.35 &21.88 &19.34 \\
+ Pretrain &1.8 &1.81 &1.8 &1.8 &1.34 &0.46 &34.33 &1.88 &1.88 &1.88 &1.88 &1.58 &0.3 &18.99 &15.34 \\
\bottomrule
\end{tabular}}
\caption{meta experiment results w.r.t ppl on GSM8K. $\mathcal{D}_\text{seen}$ denotes the sampled set from GSM8K used for training, while $\mathcal{D}_\text{unseen}$ represents the sampled unseen set for the model used for evaluation. $\mathcal{D}_\text{xxx-ref}$ stands for the synthesized referenced datasets.}
\label{tab: Meta-G-ppl}
\end{table}
%Please add the following packages if necessary:
%\usepackage{booktabs, multirow} % for borders and merged ranges
%\usepackage{soul}% for underlines
%\usepackage[table]{xcolor} % for cell colors
%\usepackage{changepage,threeparttable} % for wide tables
%If the table is too wide, replace \begin{table}[!htp]...\end{table} with
%\begin{adjustwidth}{-2.5 cm}{-2.5 cm}\centering\begin{threeparttable}[!htb]...\end{threeparttable}\end{adjustwidth}
\begin{table}[!htp]\centering
\scriptsize
\scalebox{0.65}{
\begin{tabular}{lrrrrrrr|rrrrrrr|r}
\toprule
Model & $\mathcal{D}_{\text{seen-ref}_1}$ & $\mathcal{D}_{\text{seen-ref}_2}$ & $\mathcal{D}_{\text{seen-ref}_3}$ & $\overline{\mathcal{D}_{\text{seen-ref}_i}}$ & $\mathcal{D}_{\text{seen}}$ & $\Delta_{\text{seen}}$ & $\delta_{\text{seen}}$ & 
$\mathcal{D}_{\text{unseen-ref}_1}$ & $\mathcal{D}_{\text{unseen-ref}_2}$ & $\mathcal{D}_{\text{unseen-ref}_3}$ & $\overline{\mathcal{D}_{\text{unseen-ref}_i}}$ & $\mathcal{D}_{\text{unseen}}$ &$\Delta_{\text{unseen}}$ & $\delta_{\text{unseen}}$ &\textbf{$\delta_{\text{seen-unseen}}$} \\\midrule
Mistral-7B &2.77 &2.68 &2.67 &2.71 &2.41 &0.3 &12.45 &2.83 &2.75 &2.76 &2.78 &2.49 &0.29 &11.65 &0.8 \\
+ SFT &2.16 &2.09 &2.08 &2.11 &1.53 &0.58 &37.91 &2.71 &2.64 &2.64 &2.66 &2.32 &0.34 &14.66 &23.25 \\
+ Pretrain &2.16 &2.09 &2.09 &2.11 &1.56 &0.55 &35.26 &2.69 &2.62 &2.63 &2.65 &2.3 &0.35 &15.22 &20.04 \\
\bottomrule
\end{tabular}}
\caption{Meta experiment results w.r.t ppl on MATH. $\mathcal{D}_\text{seen}$ denotes the sampled set from MATH used for training, while $\mathcal{D}_\text{unseen}$ represents the sampled unseen set for the model used for evaluation. $\mathcal{D}_\text{xxx-ref}$ stands for the synthesized referenced datasets.}
\label{tab: Meta-M-ppl}
\end{table}
%Please add the following packages if necessary:
%\usepackage{booktabs, multirow} % for borders and merged ranges
%\usepackage{soul}% for underlines
%\usepackage[table]{xcolor} % for cell colors
%\usepackage{changepage,threeparttable} % for wide tables
%If the table is too wide, replace \begin{table}[!htp]...\end{table} with
%\begin{adjustwidth}{-2.5 cm}{-2.5 cm}\centering\begin{threeparttable}[!htb]...\end{threeparttable}\end{adjustwidth}
\begin{table}[!htp]\centering
\scriptsize
\scalebox{0.68}{
\begin{tabular}{lrrrrrrr|rrrrrrr|r}
\toprule
Model & $\mathcal{D}_{\text{train-ref}_1}$ & $\mathcal{D}_{\text{train-ref}_2}$ & $\mathcal{D}_{\text{train-ref}_3}$ & $\overline{\mathcal{D}_{\text{train-ref}_i}}$ & $\mathcal{D}_{\text{train}}$ & $\Delta_{\text{train}}$ & $\delta_{\text{train}}$ & 
$\mathcal{D}_{\text{test-ref}_1}$ & $\mathcal{D}_{\text{test-ref}_2}$ & $\mathcal{D}_{\text{test-ref}_3}$ & $\overline{\mathcal{D}_{\text{test-ref}_i}}$ & $\mathcal{D}_{\text{test}}$ &$\Delta_{\text{test}}$ & $\delta_{\text{test}}$ &\textbf{$\delta_{\text{train-test}}$} \\\midrule
Qwen-7B &21.37 &21.85 &21.33 &21.52 &38.47 &16.95 &44.06 &20.15 &20.86 &20.20 &20.40 &22.30 &1.90 &8.52 &35.54 \\
Qwen-14B &23.79 &24.16 &23.60 &23.85 &48.17 &24.32 &50.49 &21.79 &22.14 &21.21 &21.71 &25.64 &3.93 &15.33 &35.16 \\
InternLM2-7B &17.71 &17.81 &17.67 &17.73 &34.76 &17.03 &48.99 &16.85 &16.12 &16.30 &16.42 &20.24 &3.82 &18.87 &30.12 \\
InternLM2-20B &18.29 &18.15 &17.93 &18.12 &36.39 &18.27 &50.21 &16.98 &16.48 &16.25 &16.57 &21.00 &4.43 &21.10 &29.11 \\
Aquila2-7B &12.58 &13.17 &12.86 &12.87 &28.69 &15.82 &55.14 &10.99 &11.28 &11.10 &11.12 &16.33 &5.21 &31.90 &23.24 \\
Aquila2-34B &16.99 &16.90 &17.11 &17.00 &37.18 &20.18 &54.28 &15.48 &15.21 &15.24 &15.31 &22.24 &6.93 &31.16 &23.12 \\
Qwen-1.8B &28.94 &29.11 &29.04 &29.03 &61.25 &32.22 &52.60 &27.01 &27.61 &26.85 &27.16 &38.65 &11.49 &29.73 &22.87 \\
Baichuan2-13B-Base &23.95 &23.81 &23.79 &23.85 &34.65 &10.80 &31.17 &22.26 &22.65 &22.44 &22.45 &25.90 &3.45 &13.32 &17.85 \\
ChatGLM2-6B &19.26 &19.55 &19.41 &19.41 &25.70 &6.29 &24.47 &18.51 &18.51 &18.59 &18.54 &20.79 &2.25 &10.82 &13.65 \\
Yi-34B &18.76 &18.27 &18.77 &18.60 &24.67 &6.07 &24.60 &17.92 &18.33 &17.92 &18.06 &20.77 &2.71 &13.05 &11.55 \\
InternLM-20B &16.65 &16.65 &16.65 &16.65 &23.21 &6.56 &28.26 &16.50 &15.59 &16.01 &16.03 &19.45 &3.42 &17.58 &10.68 \\
Phi-2 &17.43 &17.05 &17.23 &17.24 &20.07 &2.83 &14.10 &16.41 &16.09 &16.36 &16.29 &16.88 &0.59 &3.50 &10.60 \\
Yi-6B &16.54 &16.09 &16.51 &16.38 &20.68 &4.30 &20.79 &16.27 &16.88 &16.03 &16.39 &18.57 &2.18 &11.74 &9.05 \\
Orca-2-7B &24.59 &24.78 &24.47 &24.61 &35.03 &10.42 &29.75 &24.14 &24.70 &23.55 &24.13 &31.46 &7.33 &23.30 &6.45 \\
Phi-1.5 &15.67 &15.35 &15.78 &15.60 &17.02 &1.42 &8.34 &14.84 &14.94 &14.86 &14.88 &15.48 &0.60 &3.88 &4.46 \\
InternLM-7B &15.08 &14.93 &14.91 &14.97 &18.47 &3.50 &18.95 &14.98 &13.81 &14.60 &14.46 &17.03 &2.57 &15.09 &3.86 \\
LLaMA2-7B &14.45 &14.51 &14.55 &14.50 &15.43 &0.93 &6.03 &14.19 &15.33 &14.33 &14.62 &14.95 &0.33 &2.21 &3.82 \\
Aquila-7B &27.12 &27.11 &27.01 &27.08 &35.48 &8.40 &23.68 &25.85 &25.32 &26.41 &25.86 &32.78 &6.92 &21.11 &2.57 \\
Baichuan-7B &12.43 &12.47 &12.65 &12.52 &13.39 &0.87 &6.50 &12.71 &12.83 &12.78 &12.77 &13.42 &0.65 &4.84 &1.66 \\
ChatGLM3-6B &24.78 &24.96 &24.59 &24.78 &32.07 &7.29 &22.73 &24.18 &24.96 &24.85 &24.66 &31.28 &6.62 &21.16 &1.57 \\
LLaMA-7B &13.29 &13.14 &13.57 &13.33 &14.11 &0.78 &5.53 &13.10 &13.71 &13.03 &13.28 &13.90 &0.62 &4.46 &1.07 \\
Mistral-7B-v0.1 &15.96 &16.03 &16.03 &16.01 &17.07 &1.06 &6.21 &16.22 &16.60 &15.88 &16.23 &17.13 &0.90 &5.25 &0.96 \\
Grok-1 &15.54 &15.61 &15.73 &15.63 &16.37 &0.74 &4.54 &16.25 &16.01 &15.78 &16.01 &16.87 &0.86 &5.08 &-0.54 \\
Gemma-2B &15.87 &15.83 &16.10 &15.93 &15.83 &-0.10 &-0.63 &15.94 &16.22 &16.35 &16.17 &16.16 &-0.01 &-0.06 &-0.57 \\
Gemma-7B &17.93 &18.12 &18.49 &18.18 &18.75 &0.57 &3.04 &18.51 &18.62 &18.15 &18.43 &19.39 &0.96 &4.95 &-1.91 \\
Baichuan2-7B-Base &14.53 &14.57 &14.93 &14.68 &15.43 &0.75 &4.86 &14.57 &14.36 &14.62 &14.52 &15.60 &1.08 &6.92 &-2.06 \\
Llama-3-8B &12.72 &12.53 &12.96 &12.74 &12.98 &0.24 &1.87 &12.70 &12.67 &12.99 &12.79 &13.32 &0.53 &4.00 &-2.13 \\
InternLM2-20B-Base &12.75 &12.66 &12.65 &12.69 &13.49 &0.80 &5.93 &12.80 &12.46 &12.59 &12.62 &13.77 &1.15 &8.35 &-2.42 \\
Baichuan-13B-Base &14.59 &14.77 &14.63 &14.66 &15.23 &0.57 &3.74 &14.56 &14.72 &14.31 &14.53 &15.51 &0.98 &6.32 &-2.58 \\
DeepSeekMath-7B &16.31 &16.03 &16.59 &16.31 &17.25 &0.94 &5.45 &16.50 &16.95 &16.29 &16.58 &18.04 &1.46 &8.09 &-2.64 \\
InternLM2-7B-Base &12.16 &12.07 &12.25 &12.16 &11.97 &-0.19 &-1.59 &11.96 &12.16 &12.27 &12.13 &12.33 &0.20 &1.62 &-3.21 \\
\bottomrule
\end{tabular}}
\caption{LLMs ordered by  $\delta_{\text{train-test}}$  w.r.t 5-gram accuracy on GSM8K. $\mathcal{D}_\text{train}$ denotes the sampled set from GSM8K used for training, while $\mathcal{D}_\text{test}$ represents the sampled unseen set for the model used for evaluation. $\mathcal{D}_\text{xxx-ref}$ stands for the synthesized referenced datasets.}
\label{tab:GSM8K-5gram}
% \label{appendix-fig:gsm8k-5gram-acc}
\end{table}
%Please add the following packages if necessary:
%\usepackage{booktabs, multirow} % for borders and merged ranges
%\usepackage{soul}% for underlines
%\usepackage[table]{xcolor} % for cell colors
%\usepackage{changepage,threeparttable} % for wide tables
%If the table is too wide, replace \begin{table}[!htp]...\end{table} with
%\begin{adjustwidth}{-2.5 cm}{-2.5 cm}\centering\begin{threeparttable}[!htb]...\end{threeparttable}\end{adjustwidth}
\begin{table}[!htp]\centering
\scriptsize
\scalebox{0.68}{
\begin{tabular}{lrrrrrrr|rrrrrrr|r}
\toprule
Model & $\mathcal{D}_{\text{train-ref}_1}$ & $\mathcal{D}_{\text{train-ref}_2}$ & $\mathcal{D}_{\text{train-ref}_3}$ & $\overline{\mathcal{D}_{\text{train-ref}_i}}$ & $\mathcal{D}_{\text{train}}$ & $\Delta_{\text{train}}$ & $\delta_{\text{train}}$ & 
$\mathcal{D}_{\text{test-ref}_1}$ & $\mathcal{D}_{\text{test-ref}_2}$ & $\mathcal{D}_{\text{test-ref}_3}$ & $\overline{\mathcal{D}_{\text{test-ref}_i}}$ & $\mathcal{D}_{\text{test}}$ &$\Delta_{\text{test}}$ & $\delta_{\text{test}}$ &\textbf{$\delta_{\text{train-test}}$} \\\midrule
Qwen-7B &10.55 &10.99 &10.68 &10.74 &25.97 &15.23 &58.64 &10.30 &10.52 &9.98 &10.27 &11.99 &1.72 &14.35 &44.29 \\
Qwen-14B &11.89 &12.19 &11.89 &11.99 &37.05 &25.06 &67.64 &10.96 &11.30 &10.80 &11.02 &14.54 &3.52 &24.21 &43.43 \\
InternLM2-20B &7.90 &8.11 &8.05 &8.02 &29.64 &21.62 &72.94 &7.08 &7.07 &7.22 &7.12 &10.66 &3.54 &33.21 &39.73 \\
InternLM2-7B &7.55 &7.85 &7.78 &7.73 &27.26 &19.53 &71.64 &6.72 &6.76 &7.11 &6.86 &10.13 &3.27 &32.28 &39.36 \\
Aquila2-7B &3.21 &3.44 &3.23 &3.29 &17.96 &14.67 &81.68 &2.50 &2.84 &2.61 &2.65 &5.84 &3.19 &54.62 &27.06 \\
Aquila2-34B &6.40 &6.47 &6.37 &6.41 &26.79 &20.38 &76.07 &5.43 &5.94 &5.47 &5.61 &11.02 &5.41 &49.09 &26.98 \\
Qwen-1.8B &12.14 &12.40 &12.29 &12.28 &44.12 &31.84 &72.17 &11.72 &11.99 &11.80 &11.84 &23.40 &11.56 &49.40 &22.77 \\
Phi-2 &6.69 &6.85 &6.77 &6.77 &8.61 &1.84 &21.37 &6.75 &6.81 &7.20 &6.92 &6.97 &0.05 &0.72 &20.65 \\
ChatGLM2-6B &9.59 &9.77 &9.55 &9.64 &14.61 &4.97 &34.02 &9.43 &9.20 &9.42 &9.35 &10.87 &1.52 &13.98 &20.04 \\
Yi-34B &8.29 &8.29 &7.98 &8.19 &13.21 &5.02 &38.00 &8.39 &8.57 &8.45 &8.47 &10.54 &2.07 &19.64 &18.36 \\
Baichuan2-13B-Base &14.31 &14.29 &14.29 &14.30 &23.13 &8.83 &38.18 &13.75 &13.30 &13.71 &13.59 &17.06 &3.47 &20.34 &17.84 \\
Phi-1.5 &5.75 &5.78 &5.77 &5.77 &6.98 &1.21 &17.34 &5.67 &5.99 &6.13 &5.93 &5.96 &0.03 &0.50 &16.84 \\
InternLM-20B &6.55 &6.66 &6.57 &6.59 &12.27 &5.68 &46.29 &6.17 &6.17 &6.25 &6.20 &8.99 &2.79 &31.03 &15.26 \\
Yi-6B &6.81 &6.73 &6.72 &6.75 &9.77 &3.02 &30.91 &7.01 &7.13 &6.99 &7.04 &8.76 &1.72 &19.63 &11.28 \\
Baichuan-7B &4.57 &4.72 &4.47 &4.59 &5.75 &1.16 &20.17 &5.13 &5.08 &5.13 &5.11 &5.78 &0.67 &11.59 &8.58 \\
Orca-2-7b &12.10 &11.93 &11.83 &11.95 &21.71 &9.76 &44.96 &11.92 &11.63 &11.34 &11.63 &18.30 &6.67 &36.45 &8.51 \\
Mistral-7B-v0.1 &7.35 &7.65 &7.32 &7.44 &8.82 &1.38 &15.65 &7.72 &7.72 &7.98 &7.81 &8.49 &0.68 &8.01 &7.64 \\
Aquila-7B &14.55 &14.74 &14.75 &14.68 &22.07 &7.39 &33.48 &13.21 &13.18 &13.78 &13.39 &18.09 &4.70 &25.98 &7.50 \\
InternLM-7B &5.87 &5.64 &5.93 &5.81 &8.03 &2.22 &27.65 &5.43 &5.70 &6.13 &5.75 &7.26 &1.51 &20.80 &6.85 \\
LLaMA2-7B &6.37 &6.32 &6.31 &6.33 &7.09 &0.76 &10.72 &6.47 &6.52 &6.78 &6.59 &7.05 &0.46 &6.52 &4.20 \\
Baichuan-13B-Base &6.83 &6.65 &6.53 &6.67 &7.76 &1.09 &14.05 &7.17 &7.07 &6.97 &7.07 &7.87 &0.80 &10.17 &3.88 \\
Baichuan2-7B-Base &6.72 &6.75 &6.61 &6.69 &7.66 &0.97 &12.66 &7.07 &6.78 &6.91 &6.92 &7.61 &0.69 &9.07 &3.59 \\
LLaMA-7B &5.09 &5.24 &5.01 &5.11 &5.88 &0.77 &13.10 &4.97 &5.02 &5.25 &5.08 &5.64 &0.56 &9.93 &3.17 \\
Gemma-2B &7.07 &7.13 &6.98 &7.06 &9.19 &2.13 &23.18 &7.35 &7.54 &7.48 &7.46 &9.40 &1.94 &20.64 &2.54 \\
ChatGLM3-6B &11.17 &11.39 &11.17 &11.24 &17.80 &6.56 &36.85 &11.02 &11.55 &11.19 &11.25 &17.66 &6.41 &36.30 &0.55 \\
DeepSeekMath-7b &8.30 &8.46 &8.27 &8.34 &9.71 &1.37 &14.11 &8.49 &8.86 &8.54 &8.63 &10.02 &1.39 &13.87 &0.24 \\
Grok-1&	7.73	&8.00&	7.76	&7.83&	8.93&	1.10&	0.12&	8.32	&8.32	&8.52&	8.39	&9.28	&0.89&	0.10&	0.03 \\
Llama-3-8B&		7.99&		8.71&		8.61&		8.44&		9.35&		0.91	&	0.10&		7.66&		9.13&		8.83&		8.54&		9.46&		0.92&		0.10&		0.00 \\
Gemma-7B &8.47 &8.75 &8.66 &8.63 &9.75 &1.12 &11.49 &8.70 &9.29 &9.05 &9.01 &10.33 &1.32 &12.78 &-1.29 \\
InternLM2-7B-Base &5.10 &5.13 &5.30 &5.18 &5.19 &0.01 &0.19 &4.79 &5.38 &5.37 &5.18 &5.26 &0.08 &1.52 &-1.33 \\
InternLM2-20B-Base &5.28 &5.29 &5.43 &5.33 &5.68 &0.35 &6.16 &5.23 &5.25 &5.44 &5.31 &6.03 &0.72 &11.94 &-5.78 \\
\bottomrule
\end{tabular}}
\caption{LLMs ordered by  $\delta_{\text{train-test}}$  w.r.t 10-gram accuracy on GSM8K. $\mathcal{D}_\text{train}$ denotes the sampled set from GSM8K used for training, while $\mathcal{D}_\text{test}$ represents the sampled unseen set for the model used for evaluation. $\mathcal{D}_\text{xxx-ref}$ stands for the synthesized referenced datasets.}
\label{tab: GSM8K-10gram}
\end{table}
%Please add the following packages if necessary:
%\usepackage{booktabs, multirow} % for borders and merged ranges
%\usepackage{soul}% for underlines
%\usepackage[table]{xcolor} % for cell colors
%\usepackage{changepage,threeparttable} % for wide tables
%If the table is too wide, replace \begin{table}[!htp]...\end{table} with
%\begin{adjustwidth}{-2.5 cm}{-2.5 cm}\centering\begin{threeparttable}[!htb]...\end{threeparttable}\end{adjustwidth}
\begin{table}[!htp]\centering
\scriptsize
\scalebox{0.68}{
\begin{tabular}{lrrrrrrr|rrrrrrr|r}
\toprule
Model & $\mathcal{D}_{\text{train-ref}_1}$ & $\mathcal{D}_{\text{train-ref}_2}$ & $\mathcal{D}_{\text{train-ref}_3}$ & $\overline{\mathcal{D}_{\text{train-ref}_i}}$ & $\mathcal{D}_{\text{train}}$ & $\Delta_{\text{train}}$ & $\delta_{\text{train}}$ & 
$\mathcal{D}_{\text{test-ref}_1}$ & $\mathcal{D}_{\text{test-ref}_2}$ & $\mathcal{D}_{\text{test-ref}_3}$ & $\overline{\mathcal{D}_{\text{test-ref}_i}}$ & $\mathcal{D}_{\text{test}}$ &$\Delta_{\text{test}}$ & $\delta_{\text{test}}$ &\textbf{$\delta_{\text{train-test}}$} \\\midrule
InternLM2-20B &40.18 &41.63 &41.81 &41.21 &61.33 &20.12 &32.81 &31.81 &34.27 &33.33 &33.14 &37.80 &4.66 &12.33 &20.48 \\
Aquila2-34B &42.03 &44.41 &43.99 &43.48 &68.86 &25.38 &36.86 &32.39 &35.34 &34.95 &34.23 &41.93 &7.70 &18.36 &18.50 \\
InternLM2-7B &37.25 &39.56 &39.36 &38.72 &54.05 &15.33 &28.36 &30.80 &32.80 &33.11 &32.24 &36.10 &3.86 &10.69 &17.67 \\
Aquila2-7B &32.50 &34.24 &33.78 &33.51 &47.54 &14.03 &29.51 &25.76 &28.45 &28.35 &27.52 &32.10 &4.58 &14.27 &15.24 \\
Qwen-1.8B &32.41 &34.02 &33.78 &33.40 &47.31 &13.91 &29.40 &30.86 &32.83 &32.67 &32.12 &39.88 &7.76 &19.46 &9.94 \\
ChatGLM3-6B &33.63 &35.01 &35.31 &34.65 &38.57 &3.92 &10.16 &32.26 &34.13 &34.16 &33.52 &35.22 &1.70 &4.83 &5.33 \\
Qwen-7B &37.83 &39.83 &39.25 &38.97 &58.09 &19.12 &32.91 &36.53 &38.71 &38.13 &37.79 &53.01 &15.22 &28.71 &4.20 \\
InternLM-7B &18.34 &20.35 &20.41 &19.70 &22.90 &3.20 &13.97 &18.45 &20.15 &20.07 &19.56 &21.70 &2.14 &9.86 &4.11 \\
Gemma-7B &30.31 &31.49 &31.51 &31.10 &33.90 &2.80 &8.26 &29.85 &32.49 &32.07 &31.47 &33.03 &1.56 &4.72 &3.54 \\
Qwen-14B &42.77 &44.88 &44.37 &44.01 &66.87 &22.86 &34.19 &41.11 &43.01 &42.83 &42.32 &61.37 &19.05 &31.04 &3.15 \\
Llama-3-8B	&21.51&	23.15&	22.35&	22.34&	24.52&	2.18&	8.90	&21.45	&23.25&	22.90	&22.53&	23.93&	1.40&	5.84	&3.07 \\
Yi-6B &27.71 &28.84 &28.93 &28.49 &31.08 &2.59 &8.33 &28.81 &30.68 &30.37 &29.95 &31.62 &1.67 &5.28 &3.05 \\
Yi-34B &32.99 &34.68 &34.66 &34.11 &36.90 &2.79 &7.56 &33.77 &36.02 &35.40 &35.06 &36.81 &1.75 &4.75 &2.81 \\
Gemma-2B &26.87 &28.12 &27.69 &27.56 &30.03 &2.47 &8.23 &26.40 &28.44 &28.02 &27.62 &29.25 &1.63 &5.57 &2.66 \\
LLaMA-7B &22.13 &23.01 &22.74 &22.63 &25.14 &2.51 &9.98 &22.37 &23.93 &23.84 &23.38 &25.33 &1.95 &7.70 &2.28 \\
Orca-2-7b &25.55 &26.67 &26.87 &26.36 &28.03 &1.67 &5.96 &25.17 &26.60 &26.57 &26.11 &27.12 &1.01 &3.72 &2.24 \\
Phi-2 &25.12 &26.17 &26.02 &25.77 &26.75 &0.98 &3.66 &25.69 &26.95 &26.99 &26.54 &26.97 &0.43 &1.59 &2.07 \\
ChatGLM2-6B &20.43 &21.43 &21.32 &21.06 &22.89 &1.83 &7.99 &20.51 &22.28 &21.99 &21.59 &23.07 &1.48 &6.42 &1.57 \\
DeepSeekMath-7b &30.90 &32.30 &32.40 &31.87 &34.25 &2.38 &6.95 &30.78 &32.81 &32.88 &32.16 &34.07 &1.91 &5.61 &1.34 \\
Phi-1.5 &15.80 &16.97 &16.59 &16.45 &16.59 &0.14 &0.84 &16.08 &17.31 &17.09 &16.83 &16.84 &0.01 &0.06 &0.78 \\
InternLM-20B &27.46 &29.04 &29.21 &28.57 &30.92 &2.35 &7.60 &26.87 &28.61 &28.47 &27.98 &30.05 &2.07 &6.89 &0.71 \\
InternLM2-7B-Base &24.12 &26.12 &26.47 &25.57 &27.27 &1.70 &6.23 &24.89 &27.09 &26.65 &26.21 &27.78 &1.57 &5.65 &0.58 \\
InternLM2-20B-Base &21.36 &23.38 &23.02 &22.59 &26.09 &3.50 &13.42 &21.31 &23.65 &23.35 &22.77 &26.16 &3.39 &12.96 &0.46 \\
Grok-1 &	30.55	&32.53	&32.35	&31.81	&35.10&	3.29&	9.37	&30.43	&30.14&	33.00	&31.19&	34.28&	3.09&	9.01&	0.36 \\
Baichuan-7B &25.53 &26.89 &27.13 &26.52 &27.85 &1.33 &4.78 &26.15 &27.98 &27.45 &27.19 &28.51 &1.32 &4.63 &0.15 \\
LLaMA2-7B &22.01 &23.35 &23.38 &22.91 &24.23 &1.32 &5.45 &22.68 &23.99 &23.73 &23.47 &24.80 &1.33 &5.36 &0.09 \\
Mistral-7B-v0.1 &28.93 &30.32 &30.37 &29.87 &31.67 &1.80 &5.68 &28.74 &31.00 &30.57 &30.10 &32.08 &1.98 &6.17 &-0.49 \\
Baichuan2-7B-Base &29.83 &31.24 &31.06 &30.71 &31.76 &1.05 &3.31 &30.64 &32.34 &31.95 &31.64 &32.92 &1.28 &3.89 &-0.58 \\
Baichuan-13B-Base &28.40 &30.08 &29.86 &29.45 &30.96 &1.51 &4.88 &29.11 &30.83 &30.41 &30.12 &31.87 &1.75 &5.49 &-0.61 \\
Aquila-7B &23.49 &26.04 &25.85 &25.13 &26.98 &1.85 &6.86 &24.00 &26.56 &26.28 &25.61 &28.13 &2.52 &8.96 &-2.10 \\
Baichuan2-13B-Base &30.61 &32.67 &32.73 &32.00 &32.51 &0.51 &1.57 &31.44 &33.64 &32.79 &32.62 &33.97 &1.35 &3.97 &-2.40 \\
\bottomrule
\end{tabular}}
\caption{LLMs ordered by  $\delta_{\text{train-test}}$  w.r.t 5-gram accuracy on MATH. $\mathcal{D}_\text{train}$ denotes the sampled set from MATH used for training, while $\mathcal{D}_\text{test}$ represents the sampled unseen set for the model used for evaluation. $\mathcal{D}_\text{xxx-ref}$ stands for the synthesized referenced datasets.}
\label{tab: MATH-5gram}
\end{table}
%Please add the following packages if necessary:
%\usepackage{booktabs, multirow} % for borders and merged ranges
%\usepackage{soul}% for underlines
%\usepackage[table]{xcolor} % for cell colors
%\usepackage{changepage,threeparttable} % for wide tables
%If the table is too wide, replace \begin{table}[!htp]...\end{table} with
%\begin{adjustwidth}{-2.5 cm}{-2.5 cm}\centering\begin{threeparttable}[!htb]...\end{threeparttable}\end{adjustwidth}
\begin{table}[!htp]\centering
\scriptsize
\scalebox{0.68}{
\begin{tabular}{lrrrrrrr|rrrrrrr|r}
\toprule
Model & $\mathcal{D}_{\text{train-ref}_1}$ & $\mathcal{D}_{\text{train-ref}_2}$ & $\mathcal{D}_{\text{train-ref}_3}$ & $\overline{\mathcal{D}_{\text{train-ref}_i}}$ & $\mathcal{D}_{\text{train}}$ & $\Delta_{\text{train}}$ & $\delta_{\text{train}}$ & 
$\mathcal{D}_{\text{test-ref}_1}$ & $\mathcal{D}_{\text{test-ref}_2}$ & $\mathcal{D}_{\text{test-ref}_3}$ & $\overline{\mathcal{D}_{\text{test-ref}_i}}$ & $\mathcal{D}_{\text{test}}$ &$\Delta_{\text{test}}$ & $\delta_{\text{test}}$ &\textbf{$\delta_{\text{train-test}}$} \\\midrule
InternLM2-20B &25.99 &27.40 &27.77 &27.05 &54.57 &27.52 &50.43 &17.01 &18.86 &18.59 &18.15 &23.09 &4.94 &21.39 &29.04 \\
Aquila2-34B &27.49 &29.08 &29.29 &28.62 &64.21 &35.59 &55.43 &16.48 &18.55 &18.45 &17.83 &25.63 &7.80 &30.43 &25.00 \\
InternLM2-7B &22.79 &24.82 &24.61 &24.07 &42.07 &18.00 &42.79 &16.36 &18.03 &18.11 &17.50 &21.52 &4.02 &18.68 &24.11 \\
Aquila2-7B &14.25 &15.73 &15.78 &15.25 &28.91 &13.66 &47.25 &9.37 &10.88 &11.18 &10.48 &14.17 &3.69 &26.04 &21.21 \\
Qwen-1.8B &16.27 &17.54 &17.41 &17.07 &29.97 &12.90 &43.04 &15.10 &16.42 &15.81 &15.78 &23.32 &7.54 &32.33 &10.71 \\
Gemma-7B &11.14 &12.46 &12.41 &12.00 &14.37 &2.37 &16.49 &11.33 &12.89 &12.94 &12.39 &13.67 &1.28 &9.36 &7.13 \\
Qwen-7B &22.17 &23.72 &23.35 &23.08 &45.88 &22.80 &49.69 &20.50 &22.25 &22.16 &21.64 &38.68 &17.04 &44.05 &5.64 \\
InternLM-20B &12.07 &13.26 &13.81 &13.05 &16.25 &3.20 &19.69 &11.70 &13.01 &13.33 &12.68 &14.85 &2.17 &14.61 &5.08 \\
Yi-34B &12.12 &13.81 &13.78 &13.24 &15.85 &2.61 &16.47 &12.45 &14.77 &14.43 &13.88 &15.68 &1.80 &11.48 &4.99 \\
Phi-1.5 &4.01 &4.79 &4.86 &4.55 &5.40 &0.85 &15.74 &4.20 &5.14 &4.98 &4.77 &5.35 &0.58 &10.84 &4.90 \\
ChatGLM3-6B &18.24 &19.77 &19.63 &19.21 &22.61 &3.40 &15.04 &16.95 &18.45 &18.41 &17.94 &20.03 &2.09 &10.43 &4.61 \\
LLaMA2-7B &6.95 &7.98 &8.29 &7.74 &9.55 &1.81 &18.95 &7.15 &8.41 &8.14 &7.90 &9.25 &1.35 &14.59 &4.36 \\
Qwen-14B &28.11 &29.79 &29.59 &29.16 &59.56 &30.40 &51.04 &25.67 &27.42 &27.29 &26.79 &51.01 &24.22 &47.48 &3.56 \\
Yi-6B &9.21 &10.31 &10.40 &9.97 &11.79 &1.82 &15.44 &9.83 &11.15 &11.07 &10.68 &12.19 &1.51 &12.39 &3.05 \\
DeepSeekMath-7b &11.99 &13.61 &13.48 &13.03 &15.70 &2.67 &17.01 &12.18 &13.99 &13.54 &13.24 &15.40 &2.16 &14.03 &2.98 \\
ChatGLM2-6B &6.70 &7.22 &7.43 &7.12 &8.29 &1.17 &14.11 &6.96 &7.80 &7.81 &7.52 &8.53 &1.01 &11.84 &2.27 \\
Orca-2-7b &9.11 &9.92 &10.03 &9.69 &11.09 &1.40 &12.62 &8.48 &9.83 &9.61 &9.31 &10.39 &1.08 &10.39 &2.23 \\
Gemma-2B &9.25 &10.25 &10.19 &9.90 &11.59 &1.69 &14.58 &9.03 &10.61 &10.28 &9.97 &11.41 &1.44 &12.62 &1.96 \\
InternLM2-20B-Base &9.24 &10.68 &10.60 &10.17 &12.15 &1.98 &16.30 &9.23 &11.33 &11.07 &10.54 &12.31 &1.77 &14.38 &1.92 \\
Baichuan-13B-Base &10.27 &11.50 &11.59 &11.12 &12.77 &1.65 &12.92 &10.93 &12.20 &12.00 &11.71 &13.17 &1.46 &11.09 &1.83 \\
InternLM2-7B-Base &8.49 &9.73 &9.89 &9.37 &10.82 &1.45 &13.40 &8.87 &10.76 &10.36 &10.00 &11.33 &1.33 &11.74 &1.66 \\
Aquila-7B &8.13 &9.45 &9.60 &9.06 &11.52 &2.46 &21.35 &8.45 &9.98 &10.02 &9.48 &11.86 &2.38 &20.07 &1.28 \\
LLaMA-7B &7.09 &7.91 &8.20 &7.73 &9.15 &1.42 &15.52 &7.37 &8.44 &8.20 &8.00 &9.33 &1.33 &14.26 &1.26 \\
Mistral-7B-v0.1 &10.60 &11.65 &11.68 &11.31 &13.18 &1.87 &14.19 &10.83 &12.20 &11.97 &11.67 &13.45 &1.78 &13.23 &0.96 \\
Phi-2 &7.65 &8.58 &8.59 &8.27 &9.07 &0.80 &8.82 &8.05 &8.99 &8.90 &8.65 &9.45 &0.80 &8.47 &0.35 \\
Baichuan-7B &8.79 &9.83 &9.85 &9.49 &10.78 &1.29 &11.97 &8.99 &10.47 &9.97 &9.81 &11.13 &1.32 &11.86 &0.11 \\
Grok-1 &	11.89 &	13.63	 &13.43 &	12.98	 &15.87 &	2.89 &	0.18	 &11.79 &	13.57	 &13.29 &	12.88 &	15.35 &	2.47	 &0.16 &	0.02 \\
Llama-3-8B &	7.99 &	8.71 &	8.60 &	8.43 &	9.34 &	0.91 &	0.10	 &7.66 &	9.12 &	8.82 &	8.53 &	9.46	 &0.93	 &0.10	 &0.00 \\
InternLM-7B &7.37 &8.43 &8.47 &8.09 &9.92 &1.83 &18.45 &7.12 &8.21 &8.28 &7.87 &9.72 &1.85 &19.03 &-0.58 \\
Baichuan2-7B-Base &11.22 &12.49 &12.64 &12.12 &13.23 &1.11 &8.39 &11.83 &13.15 &13.02 &12.67 &14.00 &1.33 &9.50 &-1.11 \\
Baichuan2-13B-Base &11.98 &13.40 &13.72 &13.03 &14.09 &1.06 &7.52 &12.62 &14.17 &13.67 &13.49 &14.89 &1.40 &9.40 &-1.88 \\ \bottomrule
\end{tabular}}
\caption{LLMs ordered by  $\delta_{\text{train-test}}$  w.r.t 10-gram accuracy on MATH. $\mathcal{D}_\text{train}$ denotes the sampled set from MATH used for training, while $\mathcal{D}_\text{test}$ represents the sampled unseen set for the model used for evaluation. $\mathcal{D}_\text{xxx-ref}$ stands for the synthesized referenced datasets.}
\label{tab: MATH-10gram}
\end{table}
%Please add the following packages if necessary:
%\usepackage{booktabs, multirow} % for borders and merged ranges
%\usepackage{soul}% for underlines
%\usepackage[table]{xcolor} % for cell colors
%\usepackage{changepage,threeparttable} % for wide tables
%If the table is too wide, replace \begin{table}[!htp]...\end{table} with
%\begin{adjustwidth}{-2.5 cm}{-2.5 cm}\centering\begin{threeparttable}[!htb]...\end{threeparttable}\end{adjustwidth}
\begin{table}[!htp]\centering
\scriptsize
\scalebox{0.68}{
\begin{tabular}{lrrrrrrr|rrrrrrr|r}
\toprule
Model & $\mathcal{D}_{\text{train-ref}_1}$ & $\mathcal{D}_{\text{train-ref}_2}$ & $\mathcal{D}_{\text{train-ref}_3}$ & $\overline{\mathcal{D}_{\text{train-ref}_i}}$ & $\mathcal{D}_{\text{train}}$ & $\Delta_{\text{train}}$ & $\delta_{\text{train}}$ & 
$\mathcal{D}_{\text{test-ref}_1}$ & $\mathcal{D}_{\text{test-ref}_2}$ & $\mathcal{D}_{\text{test-ref}_3}$ & $\overline{\mathcal{D}_{\text{test-ref}_i}}$ & $\mathcal{D}_{\text{test}}$ &$\Delta_{\text{test}}$ & $\delta_{\text{test}}$ &\textbf{$\delta_{\text{train-test}}$} \\\midrule
Aquila2-7B &5.29 &5.30 &5.28 &5.29 &2.19 &3.10 &141.55 &5.97 &6.01 &5.94 &5.97 &3.59 &2.38 &66.30 &75.25 \\
InternLM2-20B &3.14 &3.15 &3.16 &3.15 &1.66 &1.49 &89.76 &3.43 &3.47 &3.43 &3.44 &2.82 &0.62 &21.99 &67.77 \\
InternLM2-7B &3.29 &3.29 &3.31 &3.30 &1.74 &1.56 &89.66 &3.60 &3.65 &3.58 &3.61 &2.89 &0.72 &24.91 &64.75 \\
Aquila2-34B &2.81 &2.81 &2.81 &2.81 &1.50 &1.31 &87.33 &3.01 &3.03 &3.00 &3.01 &2.04 &0.97 &47.55 &39.78 \\
ChatGLM2-6B &3.83 &3.87 &3.88 &3.86 &2.52 &1.34 &53.17 &4.21 &4.29 &4.18 &4.23 &3.40 &0.83 &24.41 &28.76 \\
Qwen-14B &2.98 &2.99 &3.00 &2.99 &1.84 &1.15 &62.50 &3.01 &3.10 &3.03 &3.05 &2.25 &0.80 &35.56 &26.94 \\
Qwen-7B &3.04 &3.04 &3.05 &3.04 &2.14 &0.90 &42.06 &3.21 &3.26 &3.19 &3.22 &2.77 &0.45 &16.25 &25.81 \\
Orca-2-7b &2.23 &2.23 &2.23 &2.23 &1.66 &0.57 &34.34 &2.37 &2.39 &2.38 &2.38 &2.03 &0.35 &17.24 &17.10 \\
InternLM-20B &3.34 &3.35 &3.36 &3.35 &2.50 &0.85 &34.00 &3.48 &3.53 &3.47 &3.49 &2.97 &0.52 &17.51 &16.49 \\
Baichuan2-13B-Base &2.08 &2.08 &2.08 &2.08 &1.62 &0.46 &28.40 &2.18 &2.20 &2.17 &2.18 &1.87 &0.31 &16.58 &11.82 \\
Qwen-1.8B &2.10 &2.10 &2.11 &2.10 &1.57 &0.53 &33.76 &2.21 &2.22 &2.20 &2.21 &1.79 &0.42 &23.46 &10.30 \\
Yi-34B &3.04 &3.04 &3.04 &3.04 &2.40 &0.64 &26.67 &3.11 &3.12 &3.11 &3.11 &2.67 &0.44 &16.48 &10.19 \\
Aquila-7B &2.98 &2.99 &2.99 &2.99 &2.02 &0.97 &48.02 &3.11 &3.13 &3.08 &3.11 &2.25 &0.86 &38.22 &9.80 \\
ChatGLM3-6B &2.34 &2.34 &2.35 &2.34 &1.87 &0.47 &25.13 &2.42 &2.45 &2.43 &2.43 &2.03 &0.40 &19.70 &5.43 \\
Yi-6B &3.36 &3.34 &3.36 &3.35 &2.78 &0.57 &20.50 &3.40 &3.42 &3.40 &3.41 &2.96 &0.45 &15.20 &5.30 \\
InternLM-7B &3.75 &3.77 &3.76 &3.76 &2.91 &0.85 &29.21 &3.84 &3.88 &3.81 &3.84 &3.07 &0.77 &25.08 &4.13 \\
Grok-1&	3.22&	3.23	&3.23&	3.23	&3.22&	0.01	&0.22&	3.21&	3.23&	3.21&	3.21&	3.19&	0.02&	0.61&	-0.39 \\
LLaMA2-7B &3.62 &3.63 &3.63 &3.63 &3.59 &0.04 &1.11 &3.61 &3.63 &3.62 &3.62 &3.57 &0.05 &1.40 &-0.29 \\
Mistral-7B-v0.1 &3.23 &3.23 &3.22 &3.23 &3.22 &0.01 &0.31 &3.23 &3.22 &3.22 &3.22 &3.19 &0.03 &0.94 &-0.63 \\
Gemma-7B &3.10 &3.10 &3.10 &3.10 &2.97 &0.13 &4.38 &3.09 &3.12 &3.09 &3.10 &2.95 &0.15 &5.08 &-0.70 \\
InternLM2-7B-Base &4.49 &4.50 &4.50 &4.50 &4.62 &-0.12 &-2.60 &4.43 &4.47 &4.43 &4.44 &4.52 &-0.08 &-1.77 &-0.83 \\
LLaMA-7B &3.70 &3.71 &3.71 &3.71 &3.63 &0.08 &2.20 &3.72 &3.72 &3.73 &3.72 &3.61 &0.11 &3.05 &-0.85 \\
Llama-3-8B&	4.22	&4.23&	4.22&	4.22&	4.06&	0.16&	3.94	&4.18	&4.19	&4.17	&4.18&	3.99&	0.19&	4.82	&-0.88\\
Baichuan-13B-Base &3.42 &3.43 &3.42 &3.42 &3.38 &0.04 &1.18 &3.42 &3.43 &3.41 &3.42 &3.35 &0.07 &2.09 &-0.91 \\
DeepSeekMath-7b &3.25 &3.26 &3.25 &3.25 &3.17 &0.08 &2.52 &3.25 &3.27 &3.26 &3.26 &3.15 &0.11 &3.49 &-0.97 \\
Baichuan-7B &3.78 &3.79 &3.79 &3.79 &3.76 &0.03 &0.80 &3.78 &3.79 &3.79 &3.79 &3.72 &0.07 &1.88 &-1.08 \\
InternLM2-20B-Base &4.08 &4.10 &4.09 &4.09 &4.21 &-0.12 &-2.85 &4.05 &4.08 &4.05 &4.06 &4.13 &-0.07 &-1.69 &-1.16 \\
Baichuan2-7B-Base &3.28 &3.28 &3.29 &3.28 &3.28 &0.00 &0.00 &3.29 &3.29 &3.28 &3.29 &3.25 &0.04 &1.23 &-1.23 \\
Gemma-2B &3.53 &3.53 &3.54 &3.53 &3.47 &0.06 &1.73 &3.52 &3.58 &3.51 &3.54 &3.43 &0.11 &3.21 &-1.48 \\
Phi-1.5 &3.74 &3.78 &3.76 &3.76 &4.17 &-0.41 &-9.83 &3.72 &3.78 &3.74 &3.75 &4.08 &-0.33 &-8.09 &-1.74 \\
Phi-2 &3.37 &3.39 &3.38 &3.38 &3.45 &-0.07 &-2.03 &3.37 &3.42 &3.37 &3.39 &3.39 &0.00 &0.00 &-2.03 \\
\bottomrule
\end{tabular}}
\caption{LLMs ordered by  $\delta_{\text{train-test}}$  w.r.t ppl on GSM8K. Note that the calculation of ppl is only performed on the solution part. $\mathcal{D}_\text{train}$ denotes the sampled set from GSM8K used for training, while $\mathcal{D}_\text{test}$ represents the sampled unseen set for the model used for evaluation. $\mathcal{D}_\text{xxx-ref}$ stands for the synthesized referenced datasets.}
\label{tab: GSM8K-ppl}
\end{table}
%Please add the following packages if necessary:
%\usepackage{booktabs, multirow} % for borders and merged ranges
%\usepackage{soul}% for underlines
%\usepackage[table]{xcolor} % for cell colors
%\usepackage{changepage,threeparttable} % for wide tables
%If the table is too wide, replace \begin{table}[!htp]...\end{table} with
% \begin{adjustwidth}{-2.5 cm}{-2.5 cm}\centering\begin{threeparttable}[!htb]
\begin{table}[!htp]\centering
\scriptsize
\scalebox{0.68}{
\begin{tabular}{lrrrrrrr|rrrrrrr|r}
\toprule
Model & $\mathcal{D}_{\text{train-ref}_1}$ & $\mathcal{D}_{\text{train-ref}_2}$ & $\mathcal{D}_{\text{train-ref}_3}$ & $\overline{\mathcal{D}_{\text{train-ref}_i}}$ & $\mathcal{D}_{\text{train}}$ & $\Delta_{\text{train}}$ & $\delta_{\text{train}}$ & 
$\mathcal{D}_{\text{test-ref}_1}$ & $\mathcal{D}_{\text{test-ref}_2}$ & $\mathcal{D}_{\text{test-ref}_3}$ & $\overline{\mathcal{D}_{\text{test-ref}_i}}$ & $\mathcal{D}_{\text{test}}$ &$\Delta_{\text{test}}$ & $\delta_{\text{test}}$ &\textbf{$\delta_{\text{train-test}}$} \\\midrule
Aquila2-7B &5.13 &4.79 &4.84 &4.92 &1.71 &3.21 &187.72 &7.37 &7.16 &6.98 &7.17 &5.56 &1.61 &28.96 &158.76 \\
InternLM2-20B &2.45 &2.39 &2.40 &2.41 &1.24 &1.17 &94.35 &3.08 &3.08 &3.03 &3.06 &2.51 &0.55 &21.91 &72.44 \\
Aquila2-34B &2.42 &2.30 &2.31 &2.34 &1.11 &1.23 &110.81 &2.98 &2.88 &2.86 &2.91 &2.07 &0.84 &40.58 &70.23 \\
InternLM2-7B &2.39 &2.31 &2.31 &2.34 &1.42 &0.92 &64.79 &2.85 &2.79 &2.78 &2.81 &2.35 &0.46 &19.57 &45.22 \\
Qwen-14B &2.47 &2.36 &2.36 &2.40 &1.26 &1.14 &90.48 &2.53 &2.43 &2.40 &2.45 &1.44 &1.01 &70.14 &20.34 \\
Qwen-7B &2.68 &2.56 &2.56 &2.60 &1.46 &1.14 &78.08 &2.77 &2.65 &2.63 &2.68 &1.67 &1.01 &60.48 &17.60 \\
Qwen-1.8B &2.98 &2.89 &2.89 &2.92 &1.97 &0.95 &48.22 &3.05 &2.96 &2.95 &2.99 &2.16 &0.83 &38.43 &9.79 \\
ChatGLM3-6B &2.60 &2.52 &2.52 &2.55 &1.95 &0.60 &30.77 &2.76 &2.68 &2.67 &2.70 &2.23 &0.47 &21.08 &9.69 \\
InternLM-7B &3.58 &3.46 &3.46 &3.50 &2.89 &0.61 &21.11 &3.67 &3.56 &3.54 &3.59 &3.12 &0.47 &15.06 &6.05 \\
InternLM-20B &3.00 &2.86 &2.85 &2.90 &2.35 &0.55 &23.40 &3.03 &2.91 &2.89 &2.94 &2.48 &0.46 &18.55 &4.85 \\
Grok-1	&2.77&	2.66&	2.67&	2.70&	2.31&	0.39	&17.01&	2.76	&2.67&	2.66	&2.70&	2.33	&0.36	&15.62	&1.39 \\
Orca-2-7b &3.88 &3.68 &3.69 &3.75 &3.53 &0.22 &6.23 &3.97 &3.80 &3.75 &3.84 &3.66 &0.18 &4.92 &1.31 \\
Gemma-2B &3.33 &3.20 &3.20 &3.24 &2.86 &0.38 &13.29 &3.30 &3.18 &3.16 &3.21 &2.86 &0.35 &12.24 &1.05 \\
Aquila-7B &3.63 &3.43 &3.44 &3.50 &2.88 &0.62 &21.53 &3.57 &3.40 &3.38 &3.45 &2.85 &0.60 &21.05 &0.48 \\
LLaMA2-7B &3.49 &3.38 &3.38 &3.42 &3.07 &0.35 &11.40 &3.44 &3.35 &3.33 &3.37 &3.03 &0.34 &11.22 &0.18 \\
Llama-3-8B&	3.04&	2.95&	2.94&	2.98&	2.59&	0.39	&14.99&	3.00&	2.93&	2.92	&2.95	&2.56	&0.38&	15.01	&-0.02 \\
Baichuan-7B &3.17 &3.11 &3.11 &3.13 &2.88 &0.25 &8.68 &3.14 &3.09 &3.07 &3.10 &2.85 &0.25 &8.77 &-0.09 \\
Baichuan2-7B-Base &2.91 &2.85 &2.85 &2.87 &2.60 &0.27 &10.38 &2.87 &2.83 &2.82 &2.84 &2.57 &0.27 &10.51 &-0.13 \\
InternLM2-7B-Base &3.19 &3.08 &3.08 &3.12 &2.75 &0.37 &13.45 &3.15 &3.06 &3.04 &3.08 &2.71 &0.37 &13.65 &-0.20 \\
Phi-2 &3.40 &3.32 &3.32 &3.35 &3.19 &0.16 &5.02 &3.35 &3.28 &3.27 &3.30 &3.13 &0.17 &5.43 &-0.41 \\
Phi-1.5 &5.11 &4.99 &4.99 &5.03 &4.93 &0.10 &2.03 &5.06 &4.98 &4.96 &5.00 &4.88 &0.12 &2.46 &-0.43 \\
Baichuan-13B-Base &2.92 &2.86 &2.86 &2.88 &2.64 &0.24 &9.09 &2.89 &2.83 &2.82 &2.85 &2.60 &0.25 &9.62 &-0.53 \\
Yi-6B &3.10 &3.00 &3.01 &3.04 &2.71 &0.33 &12.18 &3.06 &2.99 &2.97 &3.01 &2.67 &0.34 &12.73 &-0.55 \\
Baichuan2-13B-Base &2.81 &2.74 &2.74 &2.76 &2.52 &0.24 &9.52 &2.77 &2.71 &2.70 &2.73 &2.48 &0.25 &10.08 &-0.56 \\
Yi-34B &2.70 &2.60 &2.60 &2.63 &2.34 &0.29 &12.39 &2.67 &2.58 &2.57 &2.61 &2.31 &0.30 &12.99 &-0.60 \\
InternLM2-20B-Base &2.87 &2.77 &2.77 &2.80 &2.49 &0.31 &12.45 &2.83 &2.75 &2.73 &2.77 &2.45 &0.32 &13.06 &-0.61 \\
Mistral-7B-v0.1 &2.80 &2.72 &2.72 &2.75 &2.45 &0.30 &12.24 &2.76 &2.70 &2.69 &2.72 &2.41 &0.31 &12.86 &-0.62 \\
Gemma-7B &2.86 &2.74 &2.74 &2.78 &2.46 &0.32 &13.01 &2.82 &2.72 &2.71 &2.75 &2.42 &0.33 &13.64 &-0.63 \\
DeepSeekMath-7b &2.71 &2.61 &2.61 &2.64 &2.30 &0.34 &14.78 &2.68 &2.59 &2.58 &2.62 &2.27 &0.35 &15.42 &-0.64 \\
LLaMA-7B &3.48 &3.38 &3.38 &3.41 &3.08 &0.33 &10.71 &3.42 &3.35 &3.34 &3.37 &3.02 &0.35 &11.59 &-0.88 \\
ChatGLM2-6B &6.60 &6.52 &6.54 &6.55 &5.25 &1.30 &24.76 &6.46 &6.47 &6.45 &6.46 &5.14 &1.32 &25.68 &-0.92 \\
\bottomrule
\end{tabular}}
\caption{LLMs ordered by  $\delta_{\text{train-test}}$  w.r.t ppl on MATH. Note that the calculation of ppl is only performed on the solution part. $\mathcal{D}_\text{train}$ denotes the sampled set from MATH used for training, while $\mathcal{D}_\text{test}$ represents the sampled unseen set for the model used for evaluation. $\mathcal{D}_\text{xxx-ref}$ stands for the synthesized referenced datasets.}
\label{tab:MATH-ppl}
% \label{appendix-fig:math-ppl}
\end{table}

%Please add the following packages if necessary:
%\usepackage{booktabs, multirow} % for borders and merged ranges
%\usepackage{soul}% for underlines
%\usepackage[table]{xcolor} % for cell colors
%\usepackage{changepage,threeparttable} % for wide tables
%If the table is too wide, replace \begin{table}[!htp]...\end{table} with
%\begin{adjustwidth}{-2.5 cm}{-2.5 cm}\centering\begin{threeparttable}[!htb]...\end{threeparttable}\end{adjustwidth}
\begin{table}[!htp]\centering
\scriptsize
\scalebox{0.62}{
\begin{tabular}{lrrrrrr|rrrrrr|rrrrrr|rrrrrrr}\toprule
&\multicolumn{6}{c}{MATH-test} &\multicolumn{6}{c}{MATH-train} &\multicolumn{6}{c}{GSM8K-test} &\multicolumn{6}{c}{GSM8K-train} \\\cmidrule{2-25}
\textbf{Model/Accuracy} &\textbf{0/5} &\textbf{1/5} &\textbf{2/5} &\textbf{3/5} &\textbf{4/5} &\textbf{5/5} & \textbf{0/5} &\textbf{1/5} &\textbf{2/5} &\textbf{3/5} &\textbf{4/5} &\textbf{5/5} &\textbf{0/5} &\textbf{1/5} &\textbf{2/5} &\textbf{3/5} &\textbf{4/5} &\textbf{5/5} &\textbf{0/5} &\textbf{1/5} &\textbf{2/5} &\textbf{3/5} &\textbf{4/5} &\textbf{5/5} \\\midrule
Aquila2-34B &229 &729 &964 &708 &363 &7 &1 &18 &249 &1095 &1611 &26 &317 &579 &381 &42 &0 &0 &86 &777 &1617 &520 &0 &0 \\
Aquila2-7B &475 &1014 &924 &455 &132 &0 &104 &529 &926 &949 &486 &6 &499 &584 &215 &21 &0 &0 &330 &1223 &1276 &171 &0 &0 \\
Baichuan2-13B-Base &366 &988 &976 &547 &123 &0 &373 &1001 &962 &539 &125 &0 &234 &528 &492 &64 &1 &0 &257 &838 &1308 &592 &5 &0 \\
chatglm2-6b &865 &1131 &732 &231 &40 &1 &828 &1190 &706 &243 &33 &0 &357 &597 &321 &44 &0 &0 &561 &1222 &1027 &187 &3 &0 \\
chatglm3-6b &332 &926 &1014 &572 &156 &0 &269 &789 &1019 &696 &224 &3 &186 &452 &450 &213 &18 &0 &463 &961 &975 &547 &54 &0 \\
internlm2-20b &288 &896 &970 &646 &196 &4 &10 &126 &566 &1221 &1038 &39 &334 &619 &332 &34 &0 &0 &47 &814 &1830 &308 &1 &0 \\
internlm2-7b &306 &948 &954 &610 &180 &2 &41 &323 &807 &1119 &690 &20 &370 &594 &324 &31 &0 &0 &104 &934 &1705 &256 &1 &0 \\
Orca-2-7b &627 &1127 &872 &318 &56 &0 &580 &1096 &883 &374 &67 &0 &171 &482 &423 &225 &18 &0 &355 &855 &1055 &642 &93 &0 \\
phi-1\_5 &1227 &1172 &475 &116 &10 &0 &1188 &1218 &490 &95 &9 &0 &521 &594 &186 &17 &1 &0 &1141 &1212 &558 &89 &0 &0 \\
Qwen-1\_8B &277 &816 &938 &616 &328 &25 &150 &589 &902 &794 &498 &67 &67 &352 &538 &327 &35 &0 &24 &196 &654 &1045 &858 &223 \\
Qwen-7B &118 &444 &733 &874 &827 &4 &56 &269 &623 &994 &1046 &12 &315 &580 &381 &43 &0 &0 &159 &669 &1333 &833 &6 &0 \\
Qwen-14B &39 &272 &519 &843 &1319 &8 &13 &98 &318 &917 &1642 &12 &245 &550 &431 &93 &0 &0 &19 &327 &1173 &1319 &156 &6 \\
Yi-6B &401 &1110 &945 &463 &81 &0 &432 &1072 &939 &459 &97 &1 &426 &597 &260 &36 &0 &0 &864 &1354 &675 &107 &0 &0 \\
Yi-34B &248 &906 &1077 &604 &165 &0 &270 &847 &1081 &636 &165 &1 &361 &596 &312 &50 &0 &0 &612 &1297 &897 &191 &3 &0 \\
\bottomrule
\end{tabular}}
\caption{Statistics of suspicious leaked sample measured by Exact Match. The column labeled \texttt{k/5} represents the number of samples for which k correct predictions were made out of five starting points when forecasting 5-grams.}
\label{tab:Exact-Match-Instance-Statistics}
\end{table}
%Please add the following packages if necessary:
%\usepackage{booktabs, multirow} % for borders and merged ranges
%\usepackage{soul}% for underlines
%\usepackage[table]{xcolor} % for cell colors
%\usepackage{changepage,threeparttable} % for wide tables
%If the table is too wide, replace \begin{table}[!htp]...\end{table} with
%\begin{adjustwidth}{-2.5 cm}{-2.5 cm}\centering\begin{threeparttable}[!htb]...\end{threeparttable}\end{adjustwidth}
\begin{table}[!htp]\centering
\scriptsize
\scalebox{0.62}{
\begin{tabular}{lrrrrrr|rrrrrr|rrrrrr|rrrrrrr}\toprule
&\multicolumn{6}{c}{MATH-test} &\multicolumn{6}{c}{MATH-train} &\multicolumn{6}{c}{GSM8K-test} &\multicolumn{6}{c}{GSM8K-train} \\\cmidrule{2-25}
\textbf{Model/Accuracy} &\textbf{0/5} &\textbf{1/5} &\textbf{2/5} &\textbf{3/5} &\textbf{4/5} &\textbf{5/5} & \textbf{0/5} &\textbf{1/5} &\textbf{2/5} &\textbf{3/5} &\textbf{4/5} &\textbf{5/5} &\textbf{0/5} &\textbf{1/5} &\textbf{2/5} &\textbf{3/5} &\textbf{4/5} &\textbf{5/5} &\textbf{0/5} &\textbf{1/5} &\textbf{2/5} &\textbf{3/5} &\textbf{4/5} &\textbf{5/5} \\\midrule
Aquila2-34B &158 &638 &945 &819 &428 &12 &0 &14 &195 &1007 &1742 &42 &271 &555 &419 &74 &0 &0 &60 &648 &1598 &694 &0 &0 \\
Aquila2-7B &332 &926 &990 &565 &186 &1 &64 &436 &898 &1024 &567 &11 &465 &576 &247 &31 &0 &0 &303 &1161 &1314 &222 &0 &0 \\
Baichuan2-13B-Base &326 &948 &999 &590 &137 &0 &349 &944 &988 &572 &146 &1 &213 &521 &501 &82 &2 &0 &224 &798 &1316 &656 &6 &0 \\
chatglm2-6b &713 &1134 &828 &272 &52 &1 &705 &1148 &804 &301 &42 &0 &333 &580 &352 &54 &0 &0 &507 &1169 &1082 &239 &3 &0 \\
chatglm3-6b &275 &839 &1023 &662 &201 &0 &241 &698 &999 &772 &287 &3 &169 &434 &468 &222 &26 &0 &421 &949 &981 &577 &72 &0 \\
internlm2-20b &214 &768 &967 &739 &304 &8 &6 &92 &431 &1099 &1309 &63 &293 &606 &367 &53 &0 &0 &37 &733 &1820 &408 &2 &0 \\
internlm2-7b &247 &797 &954 &731 &264 &7 &27 &246 &683 &1104 &910 &30 &321 &585 &362 &51 &0 &0 &89 &862 &1689 &358 &2 &0 \\
Orca-2-7b &537 &1077 &941 &374 &71 &0 &502 &1038 &935 &434 &91 &0 &157 &461 &434 &238 &29 &0 &322 &827 &1057 &685 &109 &0 \\
phi-1\_5 &1089 &1216 &546 &136 &13 &0 &1079 &1244 &540 &127 &10 &0 &474 &598 &223 &23 &1 &0 &1027 &1211 &642 &119 &1 &0 \\
Qwen-1\_8B &213 &733 &963 &674 &388 &29 &121 &490 &882 &859 &565 &83 &62 &308 &553 &344 &52 &0 &18 &154 &608 &1032 &922 &266 \\
Qwen-7B &95 &389 &720 &905 &882 &9 &42 &227 &604 &1001 &1105 &21 &277 &550 &424 &68 &0 &0 &121 &555 &1344 &973 &7 &0 \\
Qwen-14B &31 &234 &500 &843 &1375 &17 &10 &82 &296 &881 &1714 &17 &206 &524 &474 &113 &2 &0 &14 &269 &1118 &1418 &175 &6 \\
Yi-6B &358 &1045 &976 &523 &98 &0 &386 &1007 &983 &493 &130 &1 &389 &603 &273 &54 &0 &0 &765 &1356 &731 &148 &0 &0 \\
Yi-34B &213 &837 &1090 &675 &185 &0 &244 &777 &1087 &683 &207 &2 &320 &598 &329 &71 &1 &0 &560 &1232 &953 &249 &6 &0 \\
\bottomrule
\end{tabular}}
\caption{Statistics of suspicious leaked sample measured by Edit Distance.  The column labeled \texttt{k/5} represents the number of samples for which k correct predictions were made out of five starting points when forecasting 5-grams. }
\label{tab:Edit-DIstance-Instance-Statistics}
\end{table}
%Please add the following packages if necessary:
%\usepackage{booktabs, multirow} % for borders and merged ranges
%\usepackage{soul}% for underlines
%\usepackage[table]{xcolor} % for cell colors
%\usepackage{changepage,threeparttable} % for wide tables
%If the table is too wide, replace \begin{table}[!htp]...\end{table} with
%\begin{adjustwidth}{-2.5 cm}{-2.5 cm}\centering\begin{threeparttable}[!htb]...\end{threeparttable}\end{adjustwidth}
\begin{table}[!htp]\centering
\scriptsize
\scalebox{0.62}{
\begin{tabular}{lrrrrrr|rrrrrr|rrrrrr|rrrrrrr}\toprule
&\multicolumn{6}{c}{MATH-test} &\multicolumn{6}{c}{MATH-train} &\multicolumn{6}{c}{GSM8K-test} &\multicolumn{6}{c}{GSM8K-train} \\\cmidrule{2-25}
\textbf{Model/Accuracy} &\textbf{0/5} &\textbf{1/5} &\textbf{2/5} &\textbf{3/5} &\textbf{4/5} &\textbf{5/5} & \textbf{0/5} &\textbf{1/5} &\textbf{2/5} &\textbf{3/5} &\textbf{4/5} &\textbf{5/5} &\textbf{0/5} &\textbf{1/5} &\textbf{2/5} &\textbf{3/5} &\textbf{4/5} &\textbf{5/5} &\textbf{0/5} &\textbf{1/5} &\textbf{2/5} &\textbf{3/5} &\textbf{4/5} &\textbf{5/5} \\\midrule
Aquila2-34B &80 &448 &884 &957 &596 &35 &0 &6 &134 &887 &1872 &101 &146 &490 &527 &155 &1 &0 &29 &446 &1505 &1013 &7 &0 \\
Aquila2-7B &204 &726 &1063 &723 &280 &4 &28 &272 &763 &1159 &748 &30 &285 &584 &378 &71 &1 &0 &150 &919 &1491 &439 &1 &0 \\
Baichuan2-13B-Base &159 &711 &1052 &805 &267 &6 &178 &699 &1048 &830 &239 &6 &144 &466 &560 &146 &3 &0 &119 &645 &1335 &885 &15 &1 \\
chatglm2-6b &394 &929 &1023 &532 &117 &5 &368 &997 &970 &547 &116 &2 &209 &514 &463 &128 &5 &0 &277 &933 &1318 &449 &23 &0 \\
chatglm3-6b &148 &639 &1026 &842 &333 &12 &112 &505 &955 &943 &464 &21 &118 &355 &504 &288 &54 &0 &265 &820 &1076 &703 &136 &0 \\
internlm2-20b &77 &448 &952 &997 &506 &20 &1 &32 &274 &964 &1587 &142 &173 &537 &501 &107 &1 &0 &22 &548 &1776 &649 &5 &0 \\
internlm2-7b &105 &498 &920 &981 &483 &13 &7 &97 &480 &1109 &1216 &91 &210 &533 &472 &104 &0 &0 &56 &661 &1651 &628 &4 &0 \\
Orca-2-7b &315 &905 &1059 &583 &137 &1 &283 &865 &1032 &641 &177 &2 &113 &385 &479 &283 &59 &0 &211 &689 &1075 &836 &188 &1 \\
phi-1\_5 &615 &1160 &836 &327 &61 &1 &624 &1165 &839 &318 &52 &2 &280 &545 &387 &96 &11 &0 &569 &1156 &938 &320 &16 &1 \\
Qwen-1\_8B &121 &539 &936 &828 &525 &51 &72 &331 &761 &962 &738 &136 &30 &224 &507 &443 &114 &1 &6 &88 &437 &1007 &1078 &384 \\
Qwen-7B &38 &259 &637 &961 &1082 &23 &12 &139 &483 &1011 &1317 &38 &172 &506 &525 &115 &1 &0 &66 &378 &1219 &1306 &30 &1 \\
Qwen-14B &17 &140 &421 &843 &1531 &48 &5 &46 &221 &804 &1871 &53 &118 &422 &555 &212 &12 &0 &4 &157 &885 &1634 &302 &18 \\
Yi-6B &190 &813 &1069 &709 &212 &7 &206 &786 &1028 &749 &224 &7 &251 &558 &420 &89 &1 &0 &499 &1214 &1004 &278 &5 &0 \\
Yi-34B &95 &579 &1106 &894 &322 &4 &117 &561 &1048 &920 &348 &6 &200 &535 &439 &145 &0 &0 &341 &1037 &1173 &435 &14 &0 \\
\bottomrule
\end{tabular}}
\caption{Statistics of suspicious leaked sample measured by ROUGE-L.  The column labeled \texttt{k/5} represents the number of samples for which k correct predictions were made out of five starting points when forecasting 5-grams.}
\label{tab:Rouge-L-Instance-Statistics}
\end{table}

%Please add the following packages if necessary:
%\usepackage{booktabs, multirow} % for borders and merged ranges
%\usepackage{soul}% for underlines
%\usepackage[table]{xcolor} % for cell colors
%\usepackage{changepage,threeparttable} % for wide tables
%If the table is too wide, replace \begin{table}[!htp]...\end{table} with
%\begin{adjustwidth}{-2.5 cm}{-2.5 cm}\centering\begin{threeparttable}[!htb]...\end{threeparttable}\end{adjustwidth}
\begin{table}[!htp]\centering
\scriptsize
\scalebox{0.8}{
\begin{tabular}{lr|lr|lr|lr}\toprule
Model &Download Date &Model &Download Date &Model &Download Date &Model &Download Date \\\midrule
LLaMA-7B &2023-09-30 &ChatGLM2-6B &2023-09-24 &InternLM2-7B-base &2024-02-09 &Yi-6B &2023-12-19 \\
LLaMA-2-7B &2023-08-13 &ChatGLM3-6B &2024-02-05 &InternLM2-7B &2024-02-09 &Yi-34B &2024-02-09 \\
LLaMA-3-7B &2024-04-20 &Qwen-1\_8B &2024-02-09 &InternLM2-20B-base &2024-02-09 &Phi-1.5 &2023-11-14 \\
Mistral-7B &2024-03-17 &Qwen-7B &2023-11-14 &InternLM2-20B &2024-02-09 &Phi-2 &2024-03-15 \\
Baichuan-7B &2023-11-14 &Qwen-14B &2023-10-29 &Aquila-7B &2024-02-09 &Gemma-2B &2024-03-05 \\
Baichuan2-7B-Base &2023-11-14 &InternLM-7B &2023-11-14 &Aquila2-7B &2024-02-09 &Gemma-7B &2024-03-05 \\
Baichuan-13B-Base &2023-11-14 &InternLM-20B &2024-02-09 &Aquila2-34B &2023-11-14 &Grok-1 &2024-04-04 \\
Baichuan2-13B-Base &2023-10-28 &DeepSeekMath-7B &2024-02-18 &Orca-2-7B &2023-12-19 & & \\
\bottomrule
\end{tabular}}
\caption{Tested LLMs with their respective download dates, representing the latest version available at the time of testing.}\label{tab: model download datas}
\end{table}

\clearpage

\subsection{Benchmark Transparency Card}
\label{sec:benchmark-transparency-card}

The motivation behind designing the Benchmark Transparency Card is to foster greater accountability and transparency in the evaluation of machine learning models. As the landscape of machine learning evolves, with increasingly powerful models being created, it has become challenging for users and stakeholders to assess a model's true performance and generalizability. Many models are claimed to be state-of-the-art based on their performance on benchmark datasets, but without clear information on how they were trained, optimized, and evaluated, these claims can be misleading. The Benchmark Transparency Card aims to remedy this by providing a standardized way to disclose critical information about the benchmarks a model was trained on, including any data processing or augmentation techniques that were used. In line with this, we strongly encourage the release of models alongside a ``Benchmark Transparency Card'', even when training details and data are not disclosed. At the very least, reporting the usage of benchmarks is essential to promote fair comparisons and the healthy development of the large language model field. This step towards greater openness will help build trust within the community and ensure that progress in the field is both measurable and verifiable.

As shown in Table~\ref{tab:benchmark-transparency-card}, the Benchmark Transparency Card we introduce includes three parts: basic model details, benchmark utilization statement, and benchmark evaluation details. 
We hope that this card will be widely adopted upon the release of models to foster the healthy development of large language models.

\begin{longtable}{p{4cm}|p{9cm}}
    \toprule
    \multicolumn{2}{c}{\textsc{\textbf{ Basic Model Details}}} \\
    \midrule
    \textbf{What's the name of the model and version?} &  \\ \midrule
    \textbf{Who created the model and on behalf of which entity?} &  \\ \midrule
    \textbf{What's the released date?} &   \\ \midrule
    \textbf{What's the timespan of training data?} &   \\ \midrule
    \textbf{What's the primary intended use?} &   \\ \midrule
    \textbf{Any other comment?} &  \\ \midrule
    \multicolumn{2}{c}{\textsc{\textbf{Benchmark Utilization Statement (1/n)}}} \\ \midrule
    \textbf{What's the name (and version) of this benchmark?} & \\ \midrule
    \textbf{Has the model trained on training, validation, or test sets from this benchmark? If any, provide the utilization details, such as the number of samples and the organization of data.} &  \\ \midrule
    \textbf{Detail any pre-processing or data augmentation techniques (such as paraphrasing, reformatting, and etc.) applied to the benchmark datasets and their potential impact on model performance.} &  \\ \midrule 
    \textbf{Whether this benchmark used for hyperparameter tuning? If any, provide the utilization details, such as the specific split.} & \\ \midrule
    \multicolumn{2}{c}{\textsc{\textbf{Benchmark Evaluation Details}}} \\ \midrule
    \textbf{List all benchmarks used to evaluate model performance} &  \\ \midrule
    \textbf{Describe the versions and any modifications of the benchmark datasets used.} &  \\ \midrule
    \textbf{Provide detailed performance scores of the model for each benchmark.} &  \\ \midrule
    \textbf{Disclose if any specific optimization for benchmark datasets, such as hyperparameter tuning or stopping conditions, were used.} &  \\ \midrule
    \textbf{Describe any special steps taken to achieve optimal performance on the benchmarks} &  \\ \midrule
    \textbf{Provide cross-validation results, if applicable, to demonstrate model generalization.} &  \\ \midrule
    \textbf{Discuss evaluations of the model's transfer learning abilities across different benchmarks.} &  \\ \midrule
    \textbf{Describe the metrics and methods used to measure performance on benchmarks. If non-standard metrics are used, provide detailed definitions and methodologies for calculation.} &  \\ \midrule
    \textbf{Indicate if related code, pre-trained models, or other resources are publicly available to ensure reproducibility and verifiability of results.} & \\ \bottomrule
\caption{The Benchmark Transparency Card we introduce}
\label{tab:benchmark-transparency-card} 
\vspace{-0.7cm}
\end{longtable}

\end{document}